\documentclass[final]{cvpr}

\usepackage{graphicx}
\usepackage{colortbl}

\usepackage{times}
\usepackage{epsfig}
\usepackage{amsmath}
\usepackage{amssymb}
\usepackage{arydshln}
\renewcommand{\vec}[1]{\boldsymbol{\textrm{#1}}}

\newcommand{\onevec}[1]{{\textbf 1}_{#1}}
\newcommand{\diag}[1]{\textrm{diag}(#1)}

\newcommand{\blue}[1]{#1}

\usepackage{color}
\usepackage{caption}

\usepackage[ruled,vlined]{algorithm2e}

\usepackage{multirow,array}
\newlength\savedwidth

\newcolumntype{I}{!{\vrule width 2pt}}

\newcommand{\colornewparts}[0]{\color{black}}
\newcommand{\argmax}{\mathop{\rm argmax}\limits}

\usepackage[pagebackref=true,breaklinks=true,colorlinks,bookmarks=false]{hyperref}

\def\cvprPaperID{2263} % *** Enter the CVPR Paper ID here
\def\confYear{CVPR 2021}

\begin{document}

%%%%%%%%% TITLE
%\title{Variational Monocular Depth Estimation for Reliability Prediction \\ based on a squared Mahalanobis-Wasserstein Distance}
\title{Variational Monocular Depth Estimation for Reliability Prediction}

\author{Noriaki Hirose,~~~~Shun Taguchi\thanks{Equal Contribution} ,~~~~Keisuke Kawano$^*$,~~~~Satoshi Koide\\
TOYOTA Central R$\&$D Labs., INC.\\
%Institution1 address\\
{\tt\small hirose@mosk.tytlabs.co.jp}
% For a paper whose authors are all at the same institution,
% omit the following lines up until the closing ``}''.
% Additional authors and addresses can be added with ``\and'',
% just like the second author.
% To save space, use either the email address or home page, not both
%\and
%Second Author\\
%Institution2\\
%First line of institution2 address\\
%{\tt\small secondauthor@i2.org}
}

\maketitle

%%%%%%%%% ABSTRACT
\begin{abstract}
Self-supervised learning for monocular depth estimation is widely investigated as an alternative to supervised learning approach, that requires a lot of ground truths. Previous works have successfully improved the accuracy of depth estimation by modifying the model structure, adding objectives, and masking dynamic objects and occluded area. However, when using such estimated depth image in applications, such as autonomous vehicles, and robots, we have to uniformly believe the estimated depth at each pixel position. This could lead to fatal errors in performing the tasks, because estimated depth at some pixels may make a bigger mistake. In this paper, we \blue{theoretically formulate a variational model for the} monocular depth estimation to predict the reliability of the estimated depth image. Based on the results, we can exclude the estimated depths with low reliability or refine them for actual use. The effectiveness of the proposed method is quantitatively and qualitatively demonstrated using the KITTI benchmark and Make3D dataset.
\end{abstract}

%%%%%%%%% BODY TEXT
\section{Introduction}
Three-dimensional point cloud information is essential for autonomous driving and robotics to recognize the surrounding environment and objects~\cite{thrun2002probabilistic,grisetti2007improved}. LiDAR and depth cameras are possible ways to obtain three-dimensional point cloud information. However, LiDAR is generally expensive, and the measured point cloud information is sparse even with multi-line LiDAR. In addition, the accuracy of depth cameras significantly deteriorates on highly reflective objects (e.g., mirrors), highly transparent objects (e.g., glass), and black objects. In addition, the performance will be poor in outdoors.
%In addition, it is known that it does not correctly work in outdoors.
%
\begin{figure}[t]
  \begin{center}
  %\hspace*{-5mm}
      \includegraphics[width=0.9\hsize]{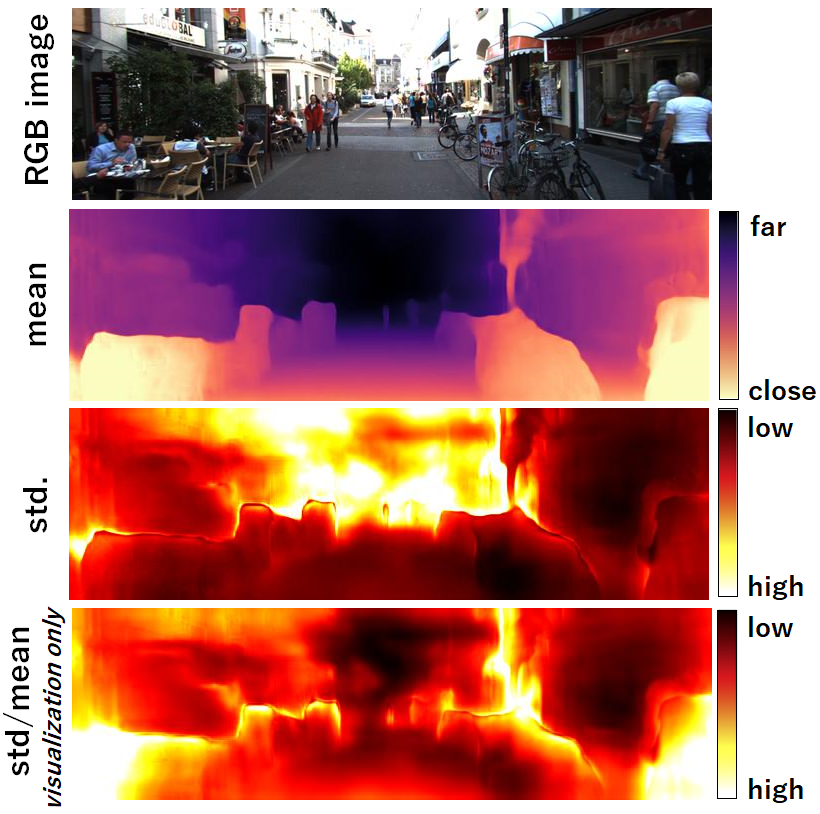}
  \end{center}
      \vspace*{-3mm}
  \caption{{\small {\bf Variational monocular depth estimation for reliability prediction.} We estimate the mean (2nd row) and standard deviation (3rd row) of depth image at each pixel from the single RGB image (1st row). {\bf Farther the area}, lower the reliability. Note that high standard deviation corresponds to low reliability. Moreover, {\bf dynamic objects}~(e.g., pedestrians) and {\bf thin objects}~(e.g., pole, bicycles) are estimated as having low reliability from the relative standard deviation (4th row).}}
  \label{f:pull_fig}
  \vspace*{-3mm}
\end{figure}

With the developments in deep learning, research on recognition and control of robots using monocular cameras has been increasing recently~\cite{NEURIPS2018_66368270,savinov2018semi,hirose2019deep}. In particular, self-supervised monocular depth estimation methods that do not require the ground truth depth have been actively studied~\cite{zhou2017unsupervised,vijayanarasimhan2017sfm}.
Instead of the ground truth, a time series of images (video), which can be captured by the robots, are used to leverage the appearance difference between two consecutive images for the depth estimation. This framework basically assumes that the environment between two consecutive frames is steady. Accordingly, lapses can occur from 1) areas distant from the camera, 2) monochromatic objects, 3) tiny and thin objects on the image, 4) dynamic objects, and 5) occluded area.

Several works attempt to overcome these weaknesses by modifying model structure~\cite{pillai2019superdepth,guizilini2019packnet,wbaf2020,fu2018deep}, adding objectives~\cite{mahjourian2018unsupervised,hirose2020plg}, and masking dynamic objects and occluded area~\cite{godard2019digging,casser2019depth,gordon2019depth}.
Although the absolute estimation error is smaller in these approaches, the estimation error remains.
Nevertheless, when using the estimated depth image for applications, such as autonomous vehicles and robots, we have to believe the estimated depth at each pixel with a uniform confidence level as no reliability information is provided.

Therefore, few studies have attempted to predict the reliability of the estimated depth image~\cite{xia2020generating,Poggi_CVPR_2020} from the depth distribution of each pixel position and measure reliability from the magnitude of the distribution. However, in Xia et al.~\cite{xia2020generating}, the ground truth depth images are needed for training. In Poggi et al.~\cite{Poggi_CVPR_2020}, a trained ensemble of multiple neural networks and/or pre-trained neural networks were used for self-teaching. Hence, the computational load and memory usage needed for online inference are high.

This paper proposes a novel variational monocular depth estimation approach to predict the reliability of the estimated depth image. It involves maximizing the likelihood of image generation using an image sequence to estimate depth distribution accurately~(Fig.~\ref{f:pull_fig}).

Our major contributions include 1) theoretical formulation of variational monocular depth estimation to predict the reliability, 2) \blue{implementation of the variational model with the Wasserstein distance whose the ground metric is the squared Mahalanobis distance}, and 3) comparative evaluation of the estimated distribution using the KITTI dataset~\cite{Geiger2013IJRR} and Make3D dataset~\cite{saxena2006learning}. 
We conducted quantitative and qualitative evaluation, including applications of the estimated distributions.
%The proposed method is evaluated quantitatively and qualitatively, including applications of the estimated distributions.
%
%
\section{Related Work}
In recent years, research on monocular depth estimation by self-supervised learning has become an important field in deep learning~\cite{fei2019geo,ummenhofer2017demon,guizilini2019packnet,patil2020don,yang2018lego,yang2017unsupervised,yang2018every,yin2018geonet,johnston2020self,zhang2020online,wang2018learning}.
The developments in spatial transformer modules have further enhanced research on monocular depth estimation~\cite{jaderberg2015spatial}.

Zhou et al.~\cite{zhou2017unsupervised}, and Vijayanarasimhan et al.~\cite{vijayanarasimhan2017sfm} were the first to apply spatial transformer modules for self-supervised monocular depth and pose estimation based on the knowledge of structure from motion (SfM).
The video clips captured by vehicles and robots were used to train the neural networks, and the depth image was estimated by minimizing the image reconstruction loss.
The studies significantly improved the depth estimation performance without the ground truth depth image.
Garg et al.~\cite{garg2016unsupervised}, and Godard et al.~\cite{godard2017unsupervised} proposed image reconstruction between stereo images to estimate the depth image, using a similar technique.
Different studies have attempted to address the lapses in existing methods.

\noindent
\textbf{Dynamic and occluded objects.} It is difficult to correctly predict the appearance of an occluded area and understand the behavior of dynamic objects from a single image.
Casser et al.~\cite{casser2019depth} proposed a motion mask using semantic segmentation to remove the dynamic object effect from the image reconstruction loss.
Gordon et al.~\cite{gordon2019depth} introduced a mobile mask to regularize the translation field for masking out the dynamic objects.
They attempt not to evaluate the occluded area based on geometric constraints.
Godard et al.~\cite{godard2019digging} presented monodepth2 with a modified image reconstruction loss by blending predicted images from previous and next images to suppress the effect of the occlusion.

\noindent
\textbf{Geometric constraint.} Several approaches attempt to penalize the geometric inconsistency between 3D point clouds projected from two different depth images. % to improve the accuracy.
However, it is difficult to penalize the geometric inconsistency as the coupling between the 3D point clouds is unknown. 
Several approaches have been proposed for estimating coupling, including iterative closest point (ICP)~\cite{mahjourian2018unsupervised}, sharing a bi-linear sampler for the image reconstruction~\cite{gordon2019depth}, the optical flow~\cite{luo2020consistent}, and the optimal transport~\cite{hirose2020plg}.
%Mahjourian et al.~\cite{mahjourian2018unsupervised} employed iterative closest point (ICP) to form a coupling between the point clouds.
%Gordon et al.~\cite{gordon2019depth} shared a bi-linear sampler for the image reconstruction to form a coupling between two depth images.
%Luo et al.~\cite{luo2020consistent} leveraged the optical flow.
%Hirose et al.~\cite{hirose2020plg} penalized the Wasserstein distance between two point clouds from different frames.

\noindent
\textbf{Post process.} Some studies refine the estimation using the past sequential images as postprocessing.
%\textbf{Post process} Following depth estimation through a neural network, some studies refine the estimation using the past sequential images.
Tiwari et al.~\cite{tiwari2020pseudo} applied SLAM to refine the estimated depth image.
Casser et al.~\cite{casser2019depth} performed fine-tuning of the pretrained neural network for inference calculations.
Unlike \cite{tiwari2020pseudo,casser2019depth}, Godard et al.~\cite{godard2017unsupervised} use both original and horizontally-flipped images as an input for depth estimation and calculate their mean to reduce artifacts.

\noindent
\textbf{Other emergent techniques.} Gordon et al.~\cite{gordon2019depth} learned a camera intrinsic parameter with depth and pose estimation for use in the wild.
Guizilini et al.~\cite{guizilini2020robust} proposed a semi-supervised learning approach that leverages small and sparse ground truth depth images, as well as packing and unpacking block~\cite{shi2016real} to maintain the spacial information.
Pillai et al.~\cite{pillai2019superdepth} demonstrated feeding a higher resolution input image to enable a more accurate estimation.

\noindent
\textbf{Probabilistic estimation.}
Pixel-wise estimation of depth distribution~\cite{xia2020generating,Poggi_CVPR_2020} is closely related to our proposed method.
%\cite{xia2020generating,Poggi_CVPR_2020} estimate distribution of depth image for each pixel and is closely related to our proposed method.
Xia et al.~\cite{xia2020generating} generates a probability density function that can express depth estimation reliability.
However, the approach uses supervised learning with the ground truth of depth images.
Poggi et al.~\cite{Poggi_CVPR_2020} show various methods to estimate the depth distribution in self-supervised learning.
However, they need to train an ensemble of multiple neural networks and/or pre-trained neural network for self-teaching.
Moreover, their performance will be limited because it depends on the pre-trained model.
Unlike these approaches, our method estimates the depth distribution without calculating multiple models and without keeping multiple parameters to reduce the computational load and memory usage during inference.
Moreover, our method can estimate more accurate distribution by learning through a loss function to maximize the likelihood of image generation.

\section{Variational Monocular Depth Estimation}

\subsection{Overview}
\begin{figure}[t]
  \begin{center}
  %\hspace*{-5mm}
      \includegraphics[width=0.99\hsize]{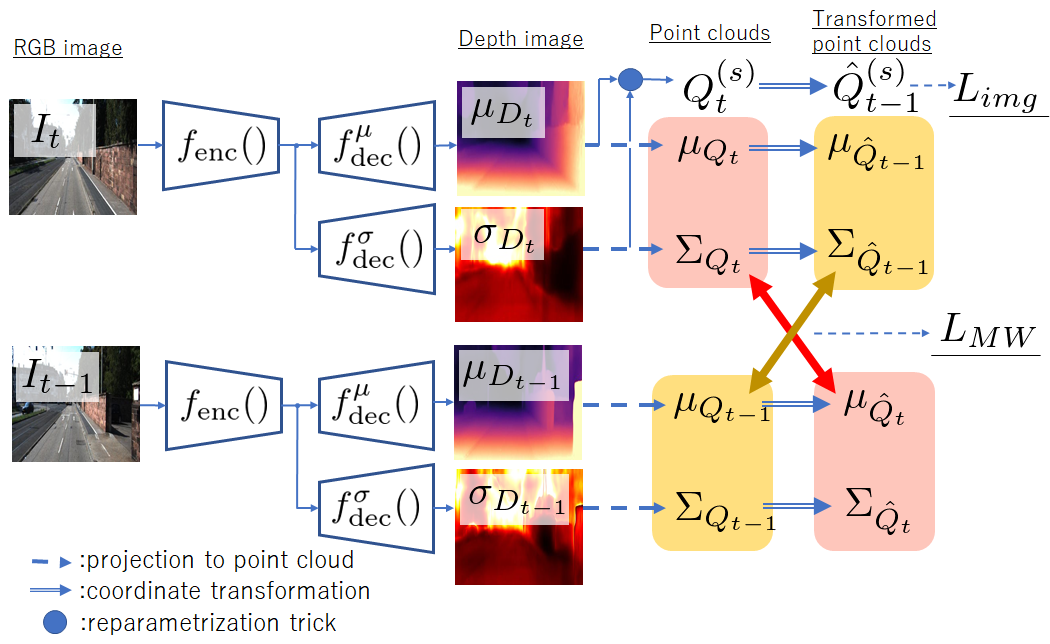}
  \end{center}
      \vspace*{-3mm}
  \caption{{\small \blue{ {\bf Overview of our proposed variational depth estimation.} We estimate the mean and standard deviation of the depth image as $\mu_{D_{t}}$, $\mu_{D_{t-1}}$ and $\sigma_{D_t}$, $\sigma_{D_{t-1}}$ from consecutive RGB images $I_t$, $I_{t-1}$. 
 Then, we project the estimated distribution on depth image coordinate into 3D space. These distribution on 3D space can be transformed into the different coordinates by using the transformation matrix. 
  We compute the image reconstruction loss $L_\text{img}$ from sampled point cloud $\hat{Q}^{(s)}_{t-1}$, and the Mahalanobis-Wasserstein loss $L_\text{MW}$ between the point cloud distributions and the transformed distributions.}}}
  \label{f:architecture}
  \vspace*{-3mm}
\end{figure}

This section introduces a probabilistic model for monocular depth estimation represented as a directed graphical model with latent variables corresponding to 3D point clouds.
Our variational monocular depth estimation is realized as a neural network-based architecture~(Fig.~\ref{f:architecture}), which will be elaborated in Sec.~\ref{sec:imp}.
The loss function for training is derived from a variational lower bound of the likelihood.

\begin{figure}[t]
    \centering
    \includegraphics[width=0.35 \linewidth]{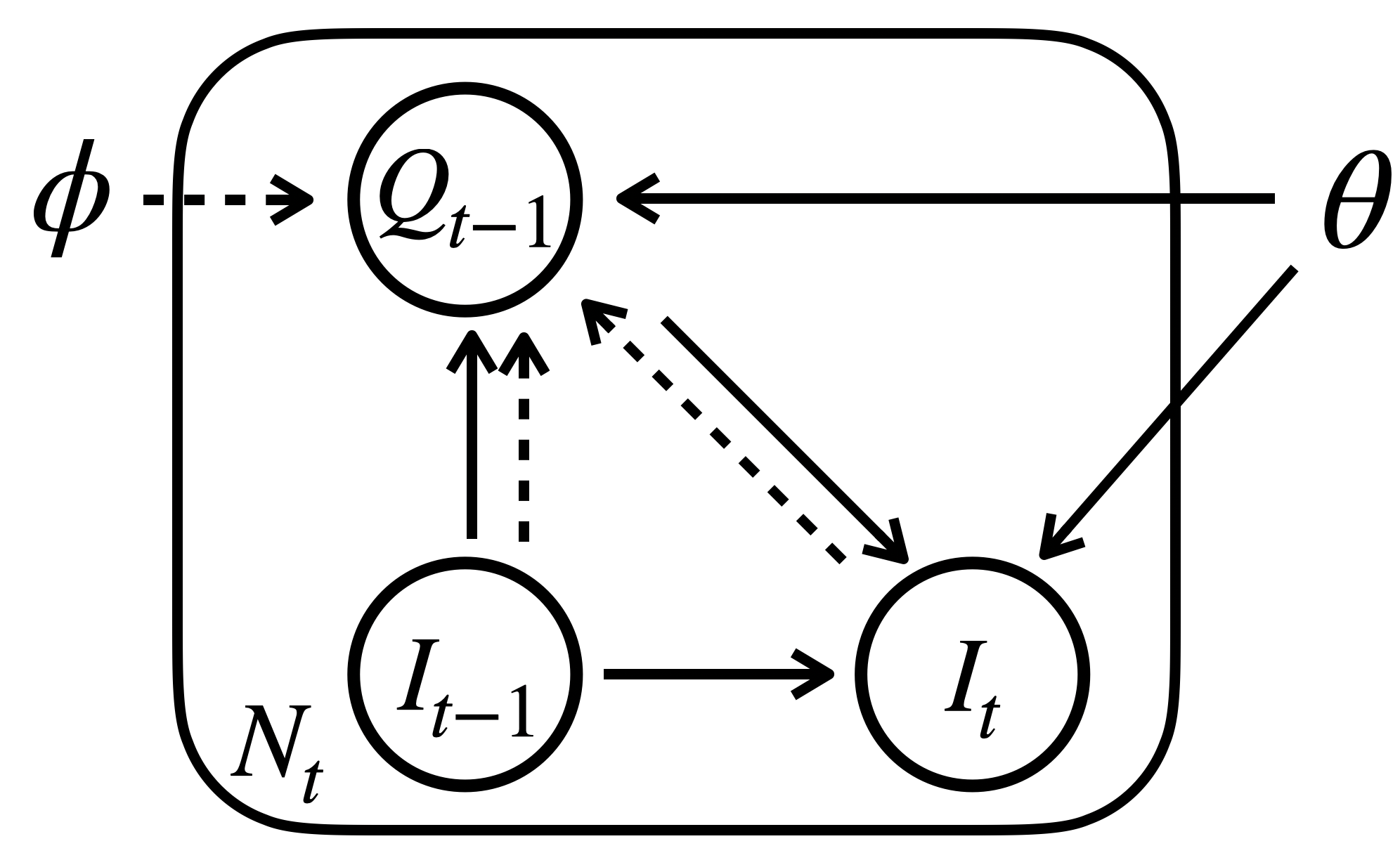}
    \caption{Directed graphical model of variational self-supervised depth estimation. The solid lines denote the generative process $p_\theta (I_t|I_{t-1})$ and $p_\theta (I_{t-1}| Q_{t-1})$, and dashed lines denote the variational approximation $q_\phi(Q_{t-1}| I_t, I_{t-1})$ to the posterior of the point cloud.}
    \label{fig:graphical-model}
\end{figure}

Let $\mathcal{I} = \{I_{0}, \cdots, I_{N_t}\}$ be the observed image sequence.
We assume that an image $I_t \in \mathbb R^{n_w\times n_h \times 3}$ obtained at time $t$ is generated by a parametric random process involving a latent random variable $\mathcal Q = \{Q_{0}, \cdots, Q_{N_t}\}$, which represents a point cloud sequence, as shown in Fig.~\ref{fig:graphical-model}.
(1) A point cloud at $t-1$ is generated from a parametric conditional distribution $p_{\theta^*}(Q_{t-1} | I_{t-1})$.
(2) An image $I_{t}$ is generated from a parametric conditional distribution $p_{\theta^*}(I_{t} | I_{t-1}, Q_{t-1})$.

The model parameters $\theta$ can be estimated via the regularized maximum likelihood inference with the marginal log likelihood of the image sequence.
\begin{align}
    \theta^* = \argmax_{\theta} \log p_\theta(\mathcal I) - \Omega(\theta),
\end{align}
where $\Omega(\theta)$ is a regularization term.
We assume that the image sequence follows Markov random process as
\begin{align}
\label{eq:loglikelihood}
   & \log p_{\theta} (\mathcal I) = \log p_\theta(I_{0}) + \sum_{t=1}^{N_t} \log p_{\theta}(I_t| I_{t-1}).
   %& \text{where}\;\; p(I_t|I_{t-1})\!=\!\!\int\! p_{\theta}(I_t| I_{t-1}, Q_{t-1}) p_{\theta}(Q_{t-1}| I_{t-1}) dQ_{t-1}\nonumber
\end{align}
Here $p(I_t|I_{t-1})\!=\!\!\int\! p_{\theta}(I_t| I_{t-1}, Q_{t-1}) p_{\theta}(Q_{t-1}| I_{t-1}) dQ_{t-1}$.
\blue{However, the integral in $p_{\theta}(I_t| I_{t-1})$ is generally intractable.}
Accordingly, we introduce a variational lower bound of  Eq.~\eqref{eq:loglikelihood} derived as follows (See supplemental for details).
\begin{align}
\label{eq:variational_bound}
\begin{split}
\log p_\theta(I_{t} | I_{t-1}) \geq & - H[q_\phi(Q_{t-1} | I_t, I_{t-1})]\\
&\hspace{-12mm}  +\mathbb{E}_{q_\phi(Q_{t-1} | I_t, I_{t-1})}\left[\log p_\theta(Q_{t-1} | I_{t-1}) \right] \\
&\hspace{-12mm} + \mathbb{E}_{q_\phi(Q_{t-1} | I_t, I_{t-1})}\left[\log p_\theta(I_t | Q_{t-1}, I_{t-1}) \right],
% & -\mathbb E_{q_\phi(Q_{t-1} | I_t, I_{t-1})}[q_\phi(Q_{t-1} | I_t, I_{t-1})],
\end{split}
\end{align}
\blue{where $H$ denotes the entropy}, and $q_\phi(Q_{t-1} | I_t, I_{t-1})$ is the variational posterior of the point cloud $Q_{t-1}$.
As shown in the next subsection, rearrangement of the variational lower bound \eqref{eq:variational_bound} conclusively leads to the following loss function for the parameter estimation:
\begin{align}
\label{eq:total-loss-function}
\mathcal L(\theta, \phi) =  L_\text{img} + \lambda_w L_\text{MW} + \Omega,
\end{align}
where $L_\text{img}$ is an image reconstruction loss that penalizes the discrepancy between the observed image $I_t$ and sampled image drawn from the posterior $p_\theta (I_t| Q_{t-1}, I_{t-1})$;
\blue{$L_\text{MW}$ penalizes inconsistency between the two point clouds $Q_t$ and $Q_{t-1}$.
Interestingly, $L_\text{MW}$ corresponds to the Wasserstein distance~\cite{peyre2019computational}, where the ground metric is the Mahalanobis distance (hereafter we call it \emph{Mahalanobis-Wasserstein distance}).}
%$L_\text{MW}$ is the Mahalanobis-Wasserstein loss that penalizes inconsistency between the two point clouds $Q_t$ and $Q_{t-1}$ using the Wasserstein distance, where the ground metric is the Mahalanobis distance.
$\lambda_w$ is a hyper parameter that controls the weight of the losses.

\subsection{Derivation of the loss function}
\blue{Here, we derive the loss function~\eqref{eq:total-loss-function} by rearranging the variational lower bound~\eqref{eq:variational_bound}.}
\\
\noindent
\textbf{Posterior distributions.}
We assume that each point $Q_{t,i}$ in the point cloud $Q_t$ follows an independent Gaussian distribution, i.e., we model the posterior $p_\theta (Q_{t-1,i} | I_{t-1})$ as;
\begin{align}
p_\theta (Q_{t-1, i} | I_{t-1}) = \prod_{j=1}^N \mathcal N (\mu_{Q_{t-1,j}}, \Sigma_{Q_{t-1,j}})^{\pi^{(i,j)}_{t-1}},
\end{align}
where $\mu_{Q_{t-1,j}}= \mu_{Q_{j}}(I_{t-1})$ and $\Sigma_{Q_{t-1,j}} = \Sigma_{Q_{j}} (I_{t-1})$ are the mean and variance of the Gaussian distribution, respectively.
$\pi^{(i,j)}_{t-1}\in \{0,1\}$ represents an \blue{unknown} correspondence between the points, that satisfies the constraints $\sum_{j=1}^N \pi^{(i,j)}_{t-1}=1, \sum_{i=1}^N \pi^{(i,j)}_{t-1} = 1$ which forces one-to-one correspondence. This $\pi$ allows to deal with the order invariance of points in the point cloud.
Note that $\mu_{Q_{t-1,i}}$ and $\Sigma_{Q_{t-1,i}}$ are implemented using the neural networks, and the parameters \blue{involved in $\mu_{Q_{t-1,i}}$, $\Sigma_{Q_{t-1,i}}$, and $\pi^{(i,j)}_{t-1}$ are included in the model parameter $\theta$.}
We also assume the variational posterior $q_\phi (Q_{t-1,j} | I_{t-1})$ as
\blue{ 
\begin{align}
q_\phi (Q_{t-1, j} | I_{t-1}, I_t) \!=\! \mathcal N (T [\mu_{Q_{t\!-\!1,j}}^\top, 1]^\top\!\!\!\!, R \Sigma_{Q_{t\!-\!1,j}}R^\top \!),
\end{align}
where $T=T_\phi (I_{t-1}, I_t)$ is a transformation matrix estimated from $I_t$ and $I_{t-1}$, which is also modeled as a neural network, %and $R$ is a rotation matrix that is part of the transformation matrix.
and $T$ is composed of a rotation matrix $R$ and a translation vector $\tau$ as $T=[R|\tau]$.
}

\noindent
\textbf{Variational lower bound.}
% The first term in the variational lower bound given in Eq.~\eqref{eq:variational_bound} is the entropy of $q_\phi(Q_{t-1} | I_t, I_{t-1})$.
\blue{The first term in the variational lower bound given in Eq.~\eqref{eq:variational_bound}, i.e., the entropy of Gaussian distribution depends on its variance as follows:}
\blue{
\begin{align}
%\begin{split}
\label{eq:first-term}
- H[q_\phi(Q_{t-1} | I_t, I_{t-1})] &= \frac{1}{2} \sum_{i=1}^N \log \left| R \Sigma_{Q_{t-1, i}} R^\top \right| + C \nonumber \\
&=\frac{1}{2} \sum_{i=1}^N \log \left|\Sigma_{Q_{t-1, i}}  \right| + C,
%\end{split}
\end{align}
where $C$ denotes a constant and $|\cdot|$ denotes the matrix determinant.
Note that the rotation matrix $R$ is orthogonal.}

Given some point clouds $\hat Q_{t-1} = \{\hat Q_{t-1}^{(1)},\cdots , \hat Q_{t-1}^{(S)}\}$ drawn from the variational posterior $q_\phi(Q_{t-1}| I_t, I_{t-1})$, the second term in Eq.~\eqref{eq:variational_bound} can be estimated as
\begin{align}
\label{eq:second-term}
\begin{split}
&\frac{1}{S} \sum_{s=1}^S  \log \prod_{i=1}^N \prod_{j=1}^N \mathcal N (\hat Q_{t-1}^{(s)} | \mu_{Q_{t-1,j}}, \Sigma_{Q_{t-1,j}})^{\pi^{(i,j)}_{t-1}} \\
%= &\frac{1}{S} \sum_{s=1}^S \sum_{i=1}^N \sum_{j=1}^N \mathcal \pi^{(i,j)}_{t-1} \left\{-\log \left(\sqrt{2\pi}^3 \sqrt{|\Sigma_{t-1, j}|} \right) \right\} - \frac{1}{2}\Delta_{t-1}^{(i,j)} \\
=&-\frac{1}{2}  \sum_{j=1}^N \log |\Sigma_{Q_{t-1,j}}| - \frac{1}{2S} \sum_{s=1}^S \sum_{i=1}^N \sum_{j=1}^N  \pi^{(i,j)}_{t-1} \Delta_{t-1}^{(i,j)}+C
% = &-\frac{1}{2}  \sum_{j=1}^N \log |\Sigma_{Q_{t-1,j}}| - \frac{1}{2S} \sum_{s=1}^S \sum_{i,j=1}^N  \pi^{(i,j)}_{t-1} \Delta_{t-1}^{(i,j)} + C
\end{split}
\end{align}
where $\Delta_{t-1}^{(i,j)} = (\hat{Q}_{t-1, i}^{(s)} - \mu_{Q_{t-1,j}}) \Sigma_{Q_{t-1,j}}^{-1} (\hat Q_{t-1, i}^{(s)} - \mu_{Q_{t-1,j}})^\top$ is the squared Mahalanobis distance.

The third term can also be estimated using the sampled point clouds as $\frac{1}{S}\sum_{s=1}^S  \log p_\theta (I_t |\hat Q_{t-1}^{(s)}, I_{t-1})$.
Assuming multivariate Gaussian distribution for the posterior distribution, i.e., $p_\theta(I_t |\hat Q_{t-1}^{(s)}, I_{t-1}) = \mathcal N (\hat{I}_t ^{(s)}, \Sigma_{I_t})$, where the mean is  $\hat{I}_t ^{(s)} = f_\text{proj}(I_{t-1}, \hat Q_{t-1}^{(s)})$ and the covariance $\Sigma_{I_t}$ is a scaled identity matrix \blue{with the scaling parameter $\lambda_{\text{img}}$} (See Sec.~\ref{sec:imp_loss} for more details of the projection function $f_\text{proj}$).
\blue{Fixing the covariance $\Sigma_{I_t}$, the third term becomes}
\begin{align}
\label{eq:third-term}
    \frac{1}{S}\sum_{s=1}^S \log \mathcal N (\hat I_t^{(s)}, \Sigma_{I_t}) %\nonumber \\ 
   % =& -\frac{1}{2S} \sum_{s=1}^S  (I_t - \hat{I}_t ^{(s)}) \Sigma_{I_t} (I_t - \hat{I}_t ^{(s)})^\top + C \nonumber \\
    = - \frac{\lambda_\text{img}}{S} \sum_{s=1}^S  \|I_t - \hat{I}_t ^{(s)}\|_2^2 + C,
\end{align}
%where
%$\lambda_\text{img}$ is the scale parameter corresponding to the scale of the covariance.
While Gaussian distribution leads to the squared error, we can utilize L1 error and SSIM~\cite{wang2004image} assuming Gibbs distribution $p_\theta(I_t | \hat Q_{t-1}^{(s)}, I_{t-1}) \propto \exp(-d_\text{img}(I_t, \hat{I}_t ^{(s)}))$, where $d_\text{img}$ is the corresponding energy function~\cite{geman1984stochastic}. %\fixme{add citation for Gibbs distribution}

\blue{Summing up the equations~\eqref{eq:first-term},~\eqref{eq:second-term}, and~\eqref{eq:third-term}, we obtain the loss function as in Eq.~\eqref{eq:total-loss-function} for the minimization problem w.r.t. $\theta$ and $\phi$, where}
\begin{align}
L_\text{img} = \frac{1}{S} \sum_{s=1}^S \sum_{t=1}^{N_t} d_\text{img}(I_t, \hat{I}_t^{(s)})
\end{align}
is the image reconstruction loss, and
\begin{align}
\label{eq:loss-geo-consistency}
    L_\text{MW} =  \min_{\pi}\frac{1}{2S} \sum_{s=1}^S \sum_{t=1}^{N_t} \sum_{i=1}^N \sum_{j=1}^N \pi^{(i,j)}_{t-1} \Delta_{t-1}^{(i,j)}
\end{align}
is the Mahalanobis-Wasserstein loss.
Note that the determinant terms in equations~\eqref{eq:first-term} and \eqref{eq:second-term} are canceled out.
% We also note that the correspondences $\pi = \{\pi^{(i,j)}_{t-1}, \forall i,j,t\}$ is relevant only to $L_\text{MW}$.
\blue{We herein include the minimization of $\pi = \{\pi^{(i,j)}_{t-1}, \forall i,j,t\}$ in Eq.~\eqref{eq:loss-geo-consistency}, because $\pi$ is relevant only to $L_\text{MW}$.}
\blue{We scale the loss so as to set $\lambda_\text{img}$ to 1 following the existing work.}

\section{Implementation}
\label{sec:imp}
In this section, we first describe how to realize the functions $\mu_{Q} (I_t) = \{\mu_{Q_{j}} (I_t)\}_{j=1,\cdots,N}$, $\Sigma_Q (I_t) =  \{\Sigma_{Q_{j}} (I_t)\}_{j=1,\cdots,N}$, and $T_\phi (I_t, I_{t-1})$ with neural networks, followed by the implementation details of the loss functions.
Fig.~\ref{f:architecture} illustrates the overview of our proposed model.

\subsection{Encoder-Decoder Model}
\label{sec:encoder-decoder-model}

We implement the point cloud distribution (i.e., $\mu_{Q}(I_t)$ and $\Sigma_{Q}(I_t)$) implicitly using the pixel-wise Gaussian distributions of the depth images.
We model both the mean and standard deviation with encoder-decoder models as follows:
\begin{align}
\mu_{D_t} = f_\text{dec}^{\mu}(f_\text{enc}(I_t)), \ \ \ 
\sigma_{D_t} = f_\text{dec}^{\sigma}(f_\text{enc}(I_t)),
\end{align}
where $I_t$ is the RGB image at the $t$-th frame, $f_\text{enc}(\cdot)$ is an encoder to extract the feature of an image, $f^{\mu}_\text{dec}(\cdot)$ is a decoder to estimate the mean of depth image $\mu_{D_t}$, and $f^{\sigma}_\text{dec}(\cdot)$ is a decoder to estimate the standard deviation of depth image $\sigma_{D_t}$.
$\mu_Q (I_t)$ and $\Sigma_Q (I_t)$ can be obtained from a linear transformation of the depth image distribution (See supplemental materials for more details).
\blue{Note that the point clouds also follow multivariate normal distribution because linearly-transformed Gaussian distribution is also Gaussian.}
The transformation matrix $T_\phi (I_t, I_{t-1})$ is modeled using the pose network as in~\cite{godard2019digging}.

\noindent
\textbf{Reparameterization trick.}
Using the reparameterization trick~\cite{kingma2013auto}, the sampled point cloud $\hat Q_{t-1}^{(s)}$ drawn from the variational posterior $q_\phi(Q_{t-1}| I_t, I_{t-1})$ can be written as
 $\hat Q_{t-1}^{(s)} = \{g_\phi (I_t, I_{t-1}, \epsilon^{(t,s,i)})\}_{i=1,\cdots,N}$, where  $\epsilon^{(t,s,i)} \sim \mathcal N (0 | 1)$.
Here, $\epsilon$ is the noise and $g_\phi$ is the differentiable transformation corresponding to the variational posterior $q_\phi(Q_{t-1}| I_t, I_{t-1})$ defined as
\blue{
\begin{align}
\hat Q_{t-1,i}^{(s)} = g_\phi  (I_t, I_{t-1}, \epsilon^{(t,s,i)}) = T  [Q_{t,i}^{(s) \top}, 1]^\top, \\
 Q_{t,i}^{(s)} = \left( \mu_{D_t}  + \epsilon^{(t,s,i)} \sigma_{D_t} \right)\cdot K^{-1} [u_i,v_i,1]^\top ,
\end{align}
}
where $K$ is the camera intrinsic parameter.
$u_i$ and $v_i$ denote the corresponding position in the image coordinate of $I_t$.
\blue{Following~\cite{kingma2013auto}, we set $S=1$ throughout this paper.}

\subsection{Loss Function.}
\label{sec:imp_loss}
\noindent
\textbf{Image reconstruction loss.}
For obtaining the projected image $\hat{I}_t = f_\text{proj}(I_{t-1} , \hat Q_{t-1})$, we basically employ the same approach as in~\cite{zhou2017unsupervised}.
By multiplying $K$ to $\hat{Q}_{t-1}$, we can obtain a mapping between $I_t$ and $I_{t-1}$ as 
% \begin{eqnarray}
% \label{eq:mapping}
 $M^{t-1}_{t} = K \cdot \hat{Q}_{t-1}$.
%\end{eqnarray}
Here $[\hat{u}, \hat{v}] = M^{t-1}_{t}(u, v)$. $\hat{u}$ and  $\hat{v}$ are the estimated pixel position corresponding to $u$ and $v$ on the image coordinate of $I_{t-1}$, respectively.
Hence, $\hat{I}_t$ can be reconstructed as
%\begin{eqnarray}
%\label{eq:sampling}
 $\hat{I}_t = f_\text{samp}(I_{t-1}, M^{t-1}_{t}  ),$
%\end{eqnarray}
where $f_\text{samp}(I_{t-1}, M^{t-1}_{t} )$ is a bi-linear sampler~\cite{jaderberg2015spatial} to sample the pixel of $I_{t-1}$ based on $M^{t-1}_{t}$.
For handling the occluded area and dynamic objects, the same image reconstruction loss function as in monodepth2~\cite{godard2019digging} is employed for the energy function $d_\text{img}(I_t, \hat{I}_t)$.

\noindent
\textbf{Mahalanobis-Wasserstein~loss.}
For calculating Eq.~\eqref{eq:loss-geo-consistency}, the discrete optimization problem with respect to the correspondence $\pi$ needs to be solved.
Because it is difficult to solve the discrete optimization problem for large point clouds, \blue{we herein solve the continuous relaxation under the constraints $\pi_{t-1}^{(i,j)} \in [0,1]$, $\sum_{i=1}^N\pi_{t-1}^{(i,j)} = 1/N$, and  $\sum_{j=1}^N\pi_{t-1}^{(i,j)} = 1/N$.
This relaxation corresponds} to the Wasserstein distance~\cite{peyre2019computational} with the squared Mahalanobis distance as the ground metric.
The Sinkhorn algorithm~\cite{cuturi2013sinkhorn} enables the quick approximation of the Wasserstein distance with GPUs.
We also employ a sampling technique to reduce the computational cost of the Wasserstein distance.

\noindent
\textbf{Regularization terms.} We employ two regularization terms $\Omega$; smooth loss and distributive regularization loss.
Penalizing roughness of the estimated depth image can improve the accuracy of the depth estimation~\cite{godard2017unsupervised,godard2019digging}, because the environment and surrounding objects are almost smooth except at the edges between the different objects.
As in \cite{godard2017unsupervised,godard2019digging}, we evaluate edge-aware smoothness of estimated depth image $\mu_{D_t}$ with the exact same loss function $L_\text{sm}$.

The distributive regularization loss $L_\text{std}$ can suppress the magnitude of the standard deviation terms as follows.
\begin{eqnarray}
\label{eq:Jstd}
L_\text{std} = \frac{1}{n_w n_h} \sum_{u=1}^{n_w} \sum_{v=1}^{n_h} \sigma_{D_t}(u,v),
\end{eqnarray}
where $n_w$ and $n_h$ denote the image size.

\blue{
Finally, the total loss function can be written as follows.
\begin{eqnarray}
\label{eq:L}
\mathcal L(\theta, \phi) = L_\text{img} + \lambda_{w} L_\text{MW} + \lambda_s L_\text{sm} + \lambda_{d} L_\text{std}
\end{eqnarray}
% Note that we scale the loss function so as to set $\lambda_\text{img}$ to 1 following the existing work.
In addition to the forward time direction of Markov process, we also consider the reverse time direction, and include the corresponding losses.
See supplemental materials for more details of the loss function.
}

\section{Experiment}
%
% In this section, quantitative and qualitative evaluations are performed on the public datasets to show the benefit of the proposed method.
%
\subsection{Dataset}
We employ the widely-used KITTI dataset~\cite{Geiger2013IJRR}, which includes the sequence of images with the camera intrinsic parameter and the LiDAR data only for testing.
Similar to baseline methods, we separate the KITTI raw dataset via the Eigen split~\cite{eigen2014depth} with 40,000 frames for training and 4,000 frames for validation. For testing, we use two sets of ground truth data: (1)~ground truth depth images from the raw LiDAR data~\cite{eigen2014depth}~({\bf Original}), and (2)~the improved ground truth by \cite{uhrig2017sparsity}~({\bf Improved}).
For comparison with various baseline methods, we employ the widely used 416$\times$128 and 640$\times$192 image sizes as the input.

{\colornewparts To evaluate the generalization ability of our method, we additionally employ Make3D dataset~\cite{saxena2006learning} to directly test the model, which is trained only on KITTI. Following \cite{zhou2016view,godard2017unsupervised,fei2019geo}, we use 134 RGB images for testing. When feeding the images, we crop center of RGB image to adapt the trained model on KITTI. Low resolution ground truth are resized for the evaluation, following \cite{zhou2016view,godard2017unsupervised,fei2019geo}.} 

For testing, we evaluate the estimated depth less than 80 m in KITTI and 70 m in Make3D, following previous works.
\subsection{Training}
We train $f_\text{enc}(\cdot)$, $f^{\mu}_\text{dec}(\cdot)$, $f^{\sigma}_\text{dec}(\cdot)$, and $f_\text{pose}(\cdot, \cdot)$ by minimizing the loss function Eq.~\eqref{eq:total-loss-function}.
We employ a two-step  training: (1) we first train with fixed standard deviation as $\sigma_{D_t}(u,v) = 1$ for $L_\text{MW}$; (2) we then train only $f^{\sigma}_\text{dec}(\cdot)$ with other networks fixed.
This is because the simultaneous training for both mean and standard deviation was difficult (a possible reason is that $f^{\sigma}_\text{dec}(\cdot)$ predicts the reliability level of depth image estimated by $f^{\mu}_\text{dec}(\cdot)$).

In both training steps, we use the same loss function $L$ with the same weighting values $\lambda_w$, $\lambda_s$, and $\lambda_{d}$ in \eqref{eq:L}, which are 0.3, 0.001 and 0.3, respectively.
%In each training iteration, we randomly sample 12 images from the training dataset of KITTI to make a mini-batch data.
Similar to \cite{Poggi_CVPR_2020}, similar network structure as in \cite{godard2019digging} is used for $f_\text{enc}(\cdot)$, $f^{\mu}_\text{dec}(\cdot)$, and $f_\text{pose}(\cdot, \cdot)$ to compare with the most related work~\cite{Poggi_CVPR_2020}.
For $f^{\sigma}_\text{dec}(\cdot)$, we adopt the same structure as $f^{\mu}_\text{dec}(\cdot)$, without sharing the weighting values.
Following the baselines, we load the pre-trained ResNet-16~\cite{he2016deep} into $f_\text{enc}(\cdot)$, before the initial step of the above training process.
We use ADAM~\cite{kingma2014adam} optimizer with a learning rate of $0.0001$ and the mini-batch size 12.

Following~\cite{hirose2020plg}, we uniformly sample the point clouds at a grid point on an image coordinate before feeding them into $L_\text{MW}$, due to the GPU memory limitation.
The vertical and horizontal grid point interval is decided as $n_c=16$ and $n_r=4$ for 416$\times$128 image size and $n_c=32$ and $n_r=5$ for 640$\times$192 image size to take more point clouds under the memory limitation.
In training, to cover the whole point clouds, we have random offsets for the grid point, which are less than and equal to $n_c$ and $n_r$ on the vertical and horizontal axes, as shown in the supplemental material.
\subsection{Quantitative analysis}
We perform three types of quantitative analysis using test data of the KITTI dataset~\cite{Geiger2013IJRR} and Make3D dataset~\cite{saxena2006learning}.
\subsubsection{Negative log-likelihood}
Unlike the several baselines, our method estimates the standard deviation as the reliability of the estimated depth image on each pixel position.
To evaluate the estimated distribution, we measure the negative log-likelihood~(NLL) over the ground truth depth values in the test data~\cite{van2016conditional}.
For comparison, we employ the following three baselines.

\vspace{2mm}
\noindent
\textbf{monodepth2~\cite{godard2019digging,Poggi_CVPR_2020}} As a post process, monodepth2~\cite{godard2019digging} feeds horizontally flipped image $\overrightarrow{I_t}$ to estimate the flipped depth image $\overrightarrow{\mu_{D_t}}$,
and flips it again as $\overleftarrow{\overrightarrow{\mu_{D_t}}}$
to calculate the mean between $\mu_{D_t}$ and $\overleftarrow{\overrightarrow{\mu_{D_t}}}$.
We derive the standard deviation as $|\mu_{D_t} - \overleftarrow{\overrightarrow{\mu_{D_t}}}|$ to calculate the NLL, following \cite{Poggi_CVPR_2020}.

\noindent
\textbf{Poggi et al.~\cite{Poggi_CVPR_2020}} propose various approaches to estimate the standard deviation in their original paper. We choose one of the best approaches called ``Self-Teaching'' to train the distribution through the pretrained model and can draw inferences from a single model.

\noindent
\textbf{Optimized uniform standard deviation~(Uniform opt.)}
This baseline estimates an {\colornewparts uniform} standard deviation $\sigma_{D_t}$ for each input image by minimizing NLL of ground truth.
For the mean depth, the same one as ours is used. Note that we cannot implement it online as it needs the ground truth.
\vspace{2mm}

Table~\ref{tab:matrix} shows the results on original and improved ground truth values in KITTI, {\colornewparts and on ground truth in Make3D.}
In addition to the three baselines above, we optimize the scale of ``Poggi et al.~\cite{Poggi_CVPR_2020}'' to make it more competitive {\colornewparts using ground truth like ``Uniform opt.''}.
Our method outperforms all baselines with healthy margins in KITTI, indicating that the distribution estimated by our proposed method has a higher probability to sample the ground truth values than the baseline methods. {\colornewparts Despite the differences in the domain of the input images, our method performs well on the Make3D dataset. Note that the ``Uniform opt.'' (with $^{\dagger}$ in Table~\ref{tab:matrix}) is better than ours, but it has used the ground truth from Make3D dataset.}
\begin{table}[t!]
\centering
\caption{{\small{\bf Negative log-likelihood evaluation} ($^{\star}$ indicates applying optimized scaling using the ground truth. $~{\dagger}$ indicates using optimized uniform standard deviation using the ground truth.)}}
\label{tab:matrix}
\resizebox{0.999\columnwidth}{!}{
\begin{tabular}{l|c|c|c} \hline
 Method & \multicolumn{2}{c|}{KITTI dataset} & Make3D \\ 
 & Original~\cite{eigen2014depth} & Improved~\cite{uhrig2017sparsity} & dataset \\ \hline
 monodepth2~\cite{godard2019digging,Poggi_CVPR_2020} & 792.822 & 1045.073 & 120.972\\
 Poggi et al.~\cite{Poggi_CVPR_2020} & 157.070 & 119.270 & 62.829 \\
 Our method & \bf{2.536} & \bf{2.322} & \bf{3.459} \\ \hline \hline
 \cellcolor[rgb]{0.9, 0.9, 0.9} Poggi et al.~\cite{Poggi_CVPR_2020}$^{\star}$ & \cellcolor[rgb]{0.9, 0.9, 0.9} 3.228 & \cellcolor[rgb]{0.9, 0.9, 0.9} 2.979 & \cellcolor[rgb]{0.9, 0.9, 0.9} 3.629 \\
 \cellcolor[rgb]{0.9, 0.9, 0.9} Uniform opt.$^{\dagger}$ & \cellcolor[rgb]{0.9, 0.9, 0.9} 2.912 & \cellcolor[rgb]{0.9, 0.9, 0.9} 2.763 & \cellcolor[rgb]{0.9, 0.9, 0.9} 3.058 \\ \hline
\end{tabular}
}
\end{table}

\begin{figure*}[t]
  \begin{center}
  %\hspace*{-5mm}
      \includegraphics[width=0.99\hsize]{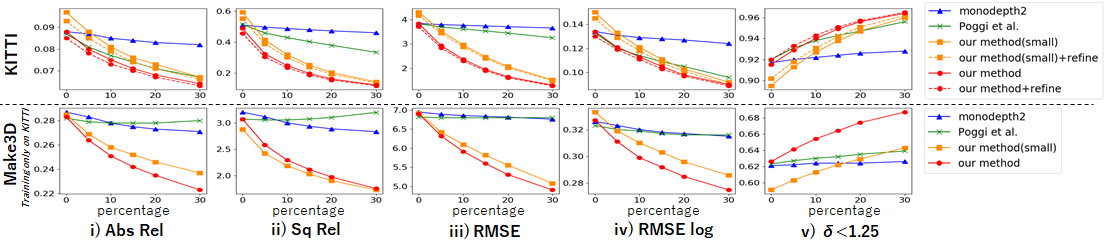}
  \end{center}
      \vspace*{-5mm}
  \caption{{\small {\bf Evaluation of outlier removal.} From left side, we show ``Abs Rel'', ``Sq Rel'', ``RMSE'', ``RMSE log'' and ``$\delta <$ 1.25'' with 0.0, 5.0, 10.0, 15.0, 20.0, and 30.0 $\%$ with outliers removed by baselines and our methods. Top graphs are for KITTI and bottom ones are for Make3D. In legend, ``small'' indicates 416$\times$128 RGB image and ``refine'' indicates refinement with the estimated standard deviation. Our methods effectively remove inaccurate estimation and obtain better results in all metrics. Our method shows healthy margins especially on ``Sq Rel'' and ``RMSE'' in KITTI. In Make3D, the generalization performance of our method is explicitly shown.}}
  \label{f:outlier}
  \vspace*{0mm}
\end{figure*}
\subsubsection{Outlier removal}
As mentioned in the introduction section, the pixels with low reliability need to be removed when using the estimated depth image.
In the second evaluation, we remove 5.0, 10.0, 15.0, 20.0, and 30.0 percentage of estimated depth from the evaluation, according to estimated $\sigma_{D_t}$.
For $N_{r}$ percentage removal, we select top-$N_{r}$ percentage pixel positions $(u,v)$ in decreasing order of the magnitude of $\sigma_{D_t}(u,v)$; these pixels are excluded from the evaluation.

Figure \ref{f:outlier} shows the results of the outlier removal for the widely used five metrics. For the most left four plots, smaller is better. For most right one, higher is better. {\colornewparts Top plots are for KITTI and bottom ones are for Make3D.} In each plot for KITTI, we show four results from our method and two baseline results. Note that ``Uniform opt.'' cannot be evaluated because the standard deviation is uniform. Our four results are a combination of different image sizes with and without refinement. {\colornewparts For Make3D, we show two results of our methods with different image size. We could not apply our refinement, because Make3D dataset does not has a time series of images.}

The performance of almost all the methods can be improved by increasing $N_r$. However, the proposed method shows superior performance compared to baseline methods. Especially, the gaps in ``Sq Rel'' and ``RMSE'' are significant in KITTI. {\colornewparts In addition, we can confirm the explicit generalization performance of our method in Make3D. The gap is significant at all removal rates.}
%the performance can drastically improve with only 5.0 $\%$ of outliers removed. 
Actual values on the plot are listed in the supplemental material.
\begin{table*}[t]
  \caption{{\small {\bf Evaluation of depth estimation by self-supervised mono supervision on an Eigen split of the KITTI dataset.} We display seven metrics from the {\bf estimated depth images less than 80 m}. For the leftmost four metrics, smaller is better; for the rightmost three metrics, higher is better. ``refine'' denotes using our proposed refinement in Sec 5.3.3. ``flip'' denotes the post process using horizontally flipped image shown in Sec 5.3.1. ``$\dagger$'' denotes using the model trained by ourselves, because monodepth2 does not open the model for 416$\times$128 image size. The bold value indicates the best result in 640$\times$192 image size. The underlined value shows the best result in 416$\times$128 image size.}}
  \vspace*{-3mm}
  \begin{center}
  \resizebox{1.9\columnwidth}{!}{
  \label{tab:ev}
  \begin{tabular}{c|lc|c|c|c|c||c|c|c} \hline
    & Method & Image size & \cellcolor[rgb]{0.9, 0.5, 0.5} Abs Rel & \cellcolor[rgb]{0.9, 0.5, 0.5} Sq Rel & \cellcolor[rgb]{0.9, 0.5, 0.5} RMSE & \cellcolor[rgb]{0.9, 0.5, 0.5} RMSE log & \cellcolor[rgb]{0.5, 0.5, 0.9} $\delta < 1.25$ & \cellcolor[rgb]{0.5, 0.5, 0.9} $\delta < 1.25^2$ & \cellcolor[rgb]{0.5, 0.5, 0.9} $\delta < 1.25^3$ \\ \hline
    \multicolumn{1}{c|}{\multirow{9}*{\rotatebox[origin=c]{90}{Original~\cite{eigen2014depth}}}} & monodepth2~\cite{godard2019digging} & 416x128 & 0.128 & 1.087 & 5.171 & 0.204 & 0.855 & 0.953 & 0.978 \\
    & monodepth2 + flip~\cite{godard2019digging} $\dagger$ & 416x128 & 0.128 & 1.108 & 5.081 & 0.203 & 0.857 & 0.954 & 0.979 \\
    & PLG-IN~\cite{hirose2020plg} & 416x128 & 0.123 & 0.920 & 4.990 & 0.201 & 0.858 & 0.953 & 0.980 \\
    & \bf{Our method} & 416x128 & 0.121 & 0.883 & 4.936 & 0.198 & 0.860 & 0.954 & 0.980 \\
    & \bf{Our method + refine} & 416x128 & \underline{0.118} & \underline{0.848} & \underline{4.845} & \underline{0.196} & \underline{0.865} & \underline{0.956} & \underline{0.981} \\
    & monodepth2~\cite{godard2019digging} & 640x192 & 0.115 & 0.903 & 4.863 & 0.193 & 0.877 & 0.959 & 0.981 \\
    & monodepth2 + flip~\cite{godard2019digging} & 640x192 & 0.112 & 0.851 & 4.754 & 0.190 & 0.881 & 0.960 & 0.981 \\
    & PLG-IN~\cite{hirose2020plg} & 640x192 & 0.114 & 0.813 & 4.705 & 0.191 & 0.874 & 0.959 & 0.981 \\
    & Poggi et al.~\cite{Poggi_CVPR_2020}  & 640x192 & 0.111 & 0.863 & 4.756 & 0.188 & \bf{0.881} & \bf{0.961} & \bf{0.982} \\
    %& Our method: & & & & & & & & \\
    & \bf{Our method} & 640x192 & 0.112 & 0.813 & 4.688 & 0.189 & 0.878 & 0.960 & \bf{0.982} \\
    & \bf{Our method + refine} & 640x192 & \bf{0.110} & \bf{0.785} & \bf{4.597} & \bf{0.187} & \bf{0.881} & \bf{0.961} & \bf{0.982} \\ \hline\hline
    \multicolumn{1}{c|}{\multirow{7}*{\rotatebox[origin=c]{90}{Improved~\cite{uhrig2017sparsity}}}} & monodepth2~\cite{godard2019digging} $\dagger$ & 416x128 & 0.108 & 0.767 & 4.343 & 0.156 & 0.890 & 0.974 & 0.991 \\
    & monodepth2 + flip~\cite{godard2019digging} $\dagger$ & 416x128 & 0.104 & 0.716 & 4.233 & 0.152 & 0.894 & 0.975 & 0.992 \\
    & \bf{Our method} & 416x128 & 0.097 & 0.592 & 4.302 & 0.150 & 0.895 & 0.975 & 0.993 \\
    & \bf{Our method + refine} & 416x128 & \underline{0.093} & \underline{0.552} & \underline{4.171} & \underline{0.145} & \underline{0.902} & \underline{0.977} & \underline{0.994} \\
    & monodepth2~\cite{godard2019digging} & 640x192 & 0.090 & 0.545 & 3.942 & 0.137 & 0.914 & 0.983 & 0.995 \\
    & monodepth2 + flip~\cite{godard2019digging} & 640x192 & 0.087 & 0.470 & 3.800 & 0.132 & 0.917 & 0.984 & 0.996 \\
    & Poggi et al.~\cite{Poggi_CVPR_2020}  & 640x192 & 0.087 & 0.514 & 3.827 & 0.133 & \bf{0.920} & 0.983 & 0.995 \\
    & \bf{Our method} & 640x192 & 0.088 & 0.488 & 3.833 & 0.134 & 0.915 & 0.983 & \bf{0.996} \\
    & \bf{Our method + refine} & 640x192 & \bf{0.085} & \bf{0.455} & \bf{3.707} & \bf{0.130} & \bf{0.920} & \bf{0.984} & \bf{0.996} \\ \hline
  \end{tabular}
  }
\end{center}
  \vspace{-5mm}
\end{table*}
\begin{figure*}[t]
  \begin{center}
  %\hspace*{-5mm}
      \includegraphics[width=0.98\hsize]{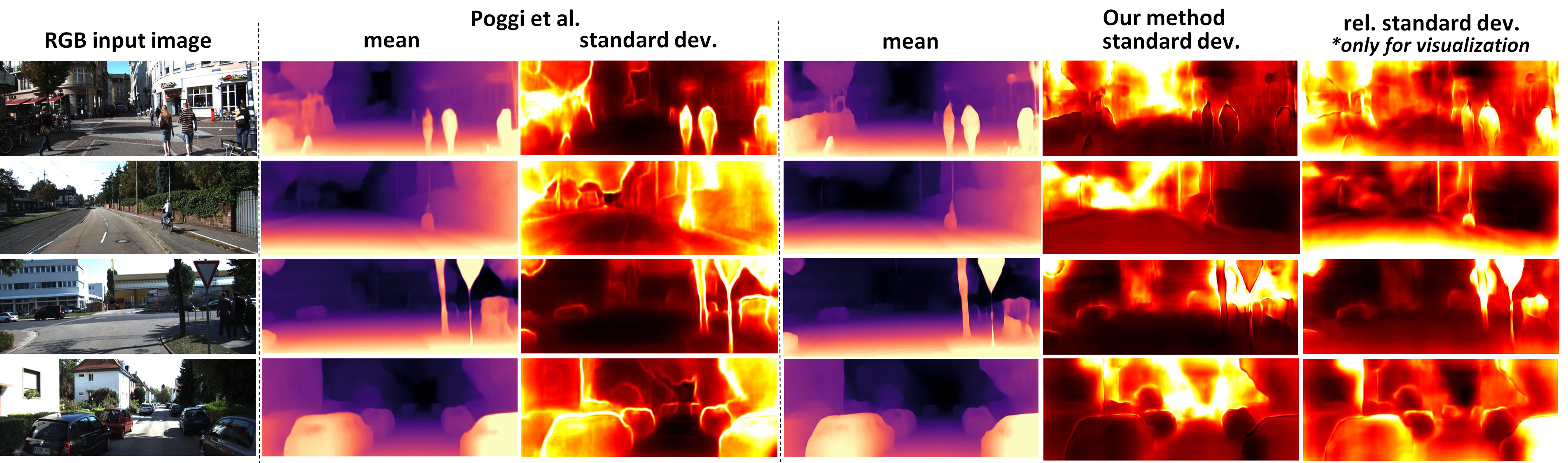}
  \end{center}
      \vspace*{-4mm}
  \caption{{\small {\bf Visualization of the proposed probabilistic depth estimation.} We display RGB image, estimated mean and standard deviation of depth image by Poggi et al.~\cite{Poggi_CVPR_2020}, and estimated mean and standard deviation of depth image from left side. In the most right side, we show the relative standard deviation to highlight low reliable pixels except farther area (visualization only). Color map of them are shown in the supplemental material.}}
  \label{f:depth}
  \vspace*{-3mm}
\end{figure*}
\begin{figure}[t]
  \begin{center}
  %\hspace*{-5mm}
      \includegraphics[width=0.87\hsize]{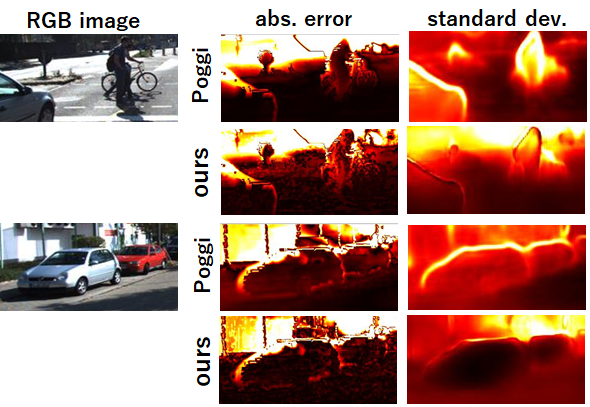}
  \end{center}
      \vspace*{-4mm}
  \caption{{\small {\bf Visualization of performance for the estimated standard deviation.} Our estimated standard deviation is visually close to the absolute error using an interpolated ground truth.}}
  \label{f:std_ev}
  \vspace*{-3mm}
\end{figure}
\begin{figure}[t]
  \begin{center}
  %\hspace*{-5mm}
      \includegraphics[width=0.87\hsize]{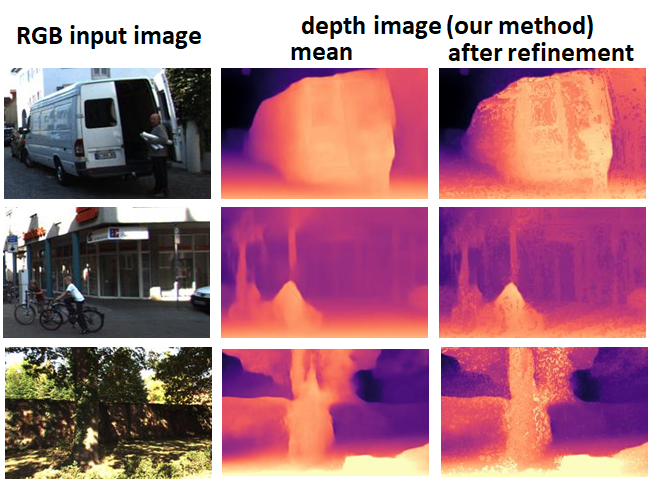}
  \end{center}
      \vspace*{-4mm}
  \caption{{\small {\bf Visualization of our refinement.} Our refinement can sharpen the depth image and decrease the artifacts.}}
  \label{f:depth_refinement}
  \vspace*{-3mm}
\end{figure}
\subsubsection{Refinement using estimated standard deviation}
\label{sec:refinement-using-estimated-std}
The final quantitative analysis is refinement using the estimated standard deviation. A stochastic optimization algorithm called the cross-entropy method~(CEM)~\cite{rubinstein2013cross,finn2017deep} is one of the approaches to find the optimal depth image from the estimated distribution. However, in monocular depth estimation, the number of states equals the number of pixels in the image leading to a vast search space; hence, applying such techniques will be difficult in an online setting.

We apply a computationally light and simple algorithm to efficiently refine the depth image using the estimated standard deviation.
\blue{
First, we sample $N_k$ depth images based on the standard deviation as follows: 
\begin{eqnarray}
    \mu_{D_t}^{k} = \mu_{D_t} + \alpha_k \cdot \sigma_{D_t}, \ \ \ k = 1, \cdots, N_k,
    \label{eq:samp}
\end{eqnarray}
where $\alpha_k$ is an equally spaced constant value between $\pm\alpha$.
Then, we reconstruct $\hat{I}^{k}_t$ as $\hat{I}^{k}_t = f_\text{proj}(I_{t-1}, \hat{Q}_{t-1}^{k})$ from the previous image $I_{t-1}$ and $\hat{Q}_{t-1}^{k}$ transformed from $\mu_{D_t}^{k}$, and calculate pixel-wise image reconstruction loss $L^k_\text{img}(u,v)$ for $N_k$ depth images.}
% First, we sample $N_{s}$ depth images at sigma points~\cite{van2004sigma} as follows: %\fixme{add citation for sigma points}
% %
% \begin{eqnarray}
%     \mu_{D_t}^{k} = \mu_{D_t} + \alpha_k \cdot \sigma_{D_t}
%     \label{eq:samp}
% \end{eqnarray}
% %
% where $\alpha_k$ is an equally spaced constant value between $\pm\alpha$. Then, we predict $\hat{I}^k_t$ by sampling the previous image $I_{t-1}$ according to Sec.~4.1 and 4.2.1, and calculate $L^k_\text{img} = 0.15\cdot|I_t - \hat{I}^k_t| + 0.85 \cdot \mbox{SSIM}(I_t, \hat{I}^k_t)$ for $N_{s}$ depth images.
% Note that the size of $L^k_\text{img}$ is same as the image size and its calculation using L1 norm and SSIM~\cite{wang2004image} is commonly used in image reconstruction loss~\cite{godard2017unsupervised,casser2019depth,gordon2019depth,godard2019digging}.
Finally, we select a depth value with the smallest $L^k_\text{img}(u,v)$ \blue{for each pixel} and obtain $\mu_{D_t}^{\star}$. To be computationally efficient for online processing, we sample only 10~$(=N_k)$ depth image from limited range~($\alpha$=0.2). On NVIDIA RTX 2080ti, our refinement algorithm needs only 2.7 ms, which is much faster than camera frame rate 100 ms in KITTI dataset.
The parameter study of $N_k$ and $\alpha$ and comparison with the post process of monodepth2 are shown in the supplemental material.

Table \ref{tab:ev} shows the results of both mean depth image $\mu_{D_t}$ and refined depth image $\mu_{D_t}^{\star}$ with some baselines based on monodepth2. The performance of our method clearly outperforms the baselines with margins in all metrics. The comparison with other state-of-the-art baselines is shown in the supplemental materials.
\subsection{Qualitative analysis}
The proposed approach is visualized in Fig.~\ref{f:depth} using KITTI test data. From left to right side, RGB input image, estimated mean and standard deviation by Poggi et al.~\cite{Poggi_CVPR_2020}, estimated mean~(=$\mu_{D_t}$), standard deviation~(= $\sigma_{D_t}$), and relative standard deviation~(= $\sigma_{D_t} / \mu_{D_t}$) are given. The estimated relative standard deviation is only for visualization to highlight low reliable pixels in not farther area. The color maps of them are shown in the supplemental material. From the estimated standard deviation, our method estimates a higher value (less reliable) in the farther area. Except for the farther area, dynamic objects (e.g., pedestrian), tiny and thin objects (e.g., pole, tree), monochromatic objects (e.g., wall, road), and edge area between objects are found to have low reliability from the relative standard deviation. In contrast, Poggi et al.~\cite{Poggi_CVPR_2020} estimate higher reliability at the farther area and monochromatic road and wall. In Fig.~\ref{f:std_ev}, we display a comparison between the estimated standard deviation and the absolute error using an interpolated ground truth. Our standard deviation is visually closer to the absolute error. {\colornewparts It means that our method is able to learn the accurate distribution.} 
%The details are shown in the supplemental material.

Figure~\ref{f:depth_refinement} shows the depth image before/after refinement. The cropped images highlight the benefits of refinement in sharpening the depth image and reducing the artifacts.
\section{Conclusion}
In this paper, we proposed a novel variational monocular depth estimation framework for reliability prediction.
For self-supervised learning, we introduced the theoretical formulation to maximize the likelihood of generation of the sequence of images.
\blue{We implemented the corresponding encoder-decoder model and loss function that involves the Mahalanobis-Wasserstein distance.}
%\blue{We implemented the probabilistic model using the encoder-decoder model trained by the loss function involving the Mahalanobis-Wasserstein distances to estimate the accurate mean and standard deviation.}

The effectiveness of our method was evaluated using KITTI dataset and Make3D dataset.
To evaluate the estimated distribution, we measured the negative-likelihood to sample the ground truth from our distribution. The benefits of our method in the outlier removal task are successfully demonstrated, further illustrating the gaps in the baseline methods. In addition, {\colornewparts the generalization performance of our method was confirmed in the evaluation, which we directly applied our model trained on KITTI to the Make3D dataset unseen during training.} 
%The benefits of our method have been demonstrated through the evaluation of the negative-log-likelihood and outlier removal. In addition, {\colornewparts the generalization performance of our method was confirmed in the evaluation, which we directly applied our model trained on KITTI to the Make3D dataset unseen during training.}
%To evaluate the estimated distribution, we measured the negative-likelihood to sample the ground truth from our distribution. The benefits of our method in removing outliers, which excludes the pixels with low reliability and offering computationally light refinement using the estimated standard deviation, are successfully demonstrated, further illustrating the gaps in the baseline methods. In addition, {\colornewparts the generalization performance of our method was confirmed in the evaluation, which we directly applied our model trained on KITTI to the Make3D dataset unseen during training.} 
%
\clearpage
{\small
\bibliographystyle{ieee_fullname}
\bibliography{egbib}
}

\clearpage
\appendix
\section{Derivation of the Variational Lower Bound}
\label{sup:variational_bound}

The variational lower bound in Eq.~\eqref{eq:variational_bound} can be derived from Eq.~\eqref{eq:loglikelihood} by introducing the variational posterior $q_\phi(Q_{t-1} | I_t, I_{t-1})$ and applying Jensen’s inequality as follows:
\begin{align}
\label{eq:variational_bound_sup}
\begin{split}
& \log p(I_{t} | I_{t-1}) \\
& \ \ = \log\! \int\! p_{\theta}(I_t| I_{t-1}, Q_{t-1}) p_{\theta}(Q_{t-1}| I_{t-1}) dQ_{t-1}, \\
& \ \ = \log \int q_\phi (Q_{t-1} | I_t, I_{t-1}) \\
& \ \ \ \ \ \ \cdot \frac{p_{\theta}(I_t| I_{t-1}, Q_{t-1}) p_{\theta}(Q_{t-1}| I_{t-1})}{q_\phi (Q_{t-1} | I_t, I_{t-1})} dQ_{t-1}, \\
& \ \ \geq \int q_\phi (Q_{t-1} | I_t, I_{t-1}) \\
& \ \ \ \ \ \ \cdot \log \frac{p_{\theta}(I_t| I_{t-1}, Q_{t-1}) p_{\theta}(Q_{t-1}| I_{t-1})}{q_\phi (Q_{t-1} | I_t, I_{t-1})} dQ_{t-1}, \\
& \ \ = - \int\! q_\phi (Q_{t-1} | I_t, I_{t-1}) \log q_\phi (Q_{t-1} | I_t, I_{t-1}) dQ_{t-1} \\
& \ \ \ \ \ \  + \int\! q_\phi (Q_{t-1} | I_t, I_{t-1}) \log p_{\theta}(Q_{t-1}| I_{t-1}) dQ_{t-1}, \\
& \ \ \ \ \ \  + \int\! q_\phi (Q_{t-1} | I_t, I_{t-1}) \log p_{\theta}(I_t| I_{t-1}, Q_{t-1}) dQ_{t-1}, \\
& \ \ = - H[q_\phi(Q_{t-1} | I_t, I_{t-1})]\\
& \ \ \ \ \ \  +\mathbb{E}_{q_\phi(Q_{t-1} | I_t, I_{t-1})}\left[\log p_\theta(Q_{t-1} | I_{t-1}) \right], \\
& \ \ \ \ \ \  + \mathbb{E}_{q_\phi(Q_{t-1} | I_t, I_{t-1})}\left[\log p_\theta(I_t | Q_{t-1}, I_{t-1}) \right], 
\end{split}
\end{align}
\section{Projection and Coordinate Transformation of Estimated Distribution}
We herein explain the transformation from the depth image distribution to the point cloud distribution shown in Section~\ref{sec:encoder-decoder-model}.
Given the pixel-wise Gaussian distribution of the depth image, the mean of a point distribution, $\mu_{Q_{t,j}}$, can be obtained as
\begin{eqnarray}
\label{eq:Q}
\mu_{Q_{t,j}} = \mu_{D_t}(u_j, v_j) \cdot K^{-1}[u_j, v_j, 1]^\top,
\end{eqnarray}
where $u_j$ and $v_j$ denote the pixel position on the image coordinates corresponding to the index of point $j$.
$K$ denotes the matrix of the camera intrinsic parameter. 

For the covariances of the point distributions, we first consider the covariance of the depth image on $[u,v,z]$ coordinates, where $z$ denotes the depth direction as follows: 
\begin{eqnarray}
\label{eq:td}
  \tilde{\Sigma}_{D_t}(u_j,v_j) = \left[
    \begin{array}{ccc}
      \sigma_u^2 & 0 & 0 \\
      0 & \sigma_v^2 & 0 \\
      0 & 0 & \sigma_{D_t}(u_j,v_j)^2
    \end{array}
  \right],
\end{eqnarray}
where $\sigma_u$ and $\sigma_v$ denote the standard deviations on the $u$ and $v$ axes, respectively.
$\sigma_{D_t}(u_j,v_j)$ is the standard deviation of the depth image at the $(u_j,v_j)$ position obtained from the encoder–-decoder model (Sec.~\ref{sec:encoder-decoder-model}).
We can formulate the covariance matrix of the point distribution on the orthogonal coordinate system as
%
%%%%%%%%%%%%%%%%%%
\begin{eqnarray}
\label{eq:Q-t}
\Sigma_{Q_{t,j}} = \Gamma_{j} \cdot \tilde{\Sigma}_{D_t} (u_j,v_j) \cdot \Gamma_{j}^\top ,
\end{eqnarray}
where $\Gamma_j=\Gamma(u_j, v_j, \mu_{D_t}(u_j, v_j))$ is the Jacobian matrix of the coordinate transformation from $[u, v, z]^\top$ to the orthogonal coordinate system, denoted as $[x, y, z]^\top$.
Here, $\Gamma$ is provided as follows:
%\begin{eqnarray}
%\label{eq:gamma}
%  \Gamma(u,v,z) = \frac{\partial[x, y, z]^\top}{\partial [u, v, z]^\top}.
%\end{eqnarray}
\begin{eqnarray}
\label{eq:gamma}
  \Gamma (u,v,z) = \frac{\partial[x, y, z]^\top}{\partial [u, v, z]^\top} = \left[
    \begin{array}{ccc}
      \frac{z}{f_x} & 0 & \frac{u-u_c}{f_x} \\
      0 & \frac{z}{f_y} & \frac{v-v_c}{f_y} \\
      0 & 0 & 1
    \end{array}
  \right],
\end{eqnarray}
where  $u_c$ and $v_c$ are center of image on $u$ and $v$ axis, and $f_x$ and $f_y$ are focal length on $x$ and $y$ axis, respectively.
$\Gamma$ is defined for the rectified image of KITTI dataset.
In our experiment, we set both $\sigma_u$, and $\sigma_v$ as a constant value, 0.5 to consider a quantization error due to lower resolution and a lens distortion.
\section{Details of the Loss Functions}
We provide the details of the loss functions shown in Section~\ref{sec:imp_loss}.
\subsection{Image reconstruction loss}
We employ the \emph{photometric reconstruction error function} involving SSIM~\cite{wang2004image} and L1 loss shown in~\cite{godard2019digging} for the energy function of the image reconstruction loss. 
\begin{align}
    & d_\text{img}(I_t, \hat I_t^{(s)}) = \frac{\beta}{2} (1-\text{SSIM}(I_t, \hat I_t^{(s)})) \\ \nonumber &\hspace{25mm} + (1-\beta) \|I_t - \hat I_t^{(s)} \|_1,
\end{align} 
where $\beta=0.85$. 
To deal with occluded areas and dynamic objects in the photometric reconstruction error function, we introduce ``per-pixel minimum reprojection loss,'' ``auto-masking stationary pixels,'' and ``multi-scale estimation'' following \cite{godard2019digging}.
\subsection{Calculation of Mahalanobis--Wasserstein Loss}
To apply Eq.~\eqref{eq:loss-geo-consistency} with the relaxed conditions as a loss function in end-to-end neural network training, we need an algorithm to compute the optimal correspondence $\pi_t^*$, where $\pi_t = [\pi^{(i,j)}_{t}]_{i,j} \in [0,1]^{N\times N}$, and the gradient with respect to the Mahalanobis distances.
We introduce an $N$-by-$N$ matrix $\Delta_t$, where $(i,j)$ element is the squared Mahalanobis distance $\Delta_t^{(i,j)}$ between the sampled point cloud $\hat Q_{t-1,j}$ drawn from the variational posterior $q_\phi (Q_{t-1,j}| I_t, I_{t-1})$ and the distribution $\mathcal{N}(\mu_{Q_t, i},\,\Sigma_{Q_t, i})$. 
Here, we can rewrite the Mahalanobis--Wasserstein loss as
\begin{align}
    L_\text{MW} =  \sum_{t=1}^{N_t} \min_{\pi_t \in \mathcal U} \left< \pi_t, \Delta_t \right>
\end{align}
where $\left< \cdot, \cdot \right>$ denotes the matrix inner product, and $\mathcal U = \{\pi |\pi \in [0,1]^{N\times N}, \sum_{i=1}^N \pi_i=\frac{1}{N}, \sum_{j=1}^N \pi_j = \frac{1}{N} \}$ is the set of valid correspondence matrices. 
This denotes the sum of the Wasserstein distances, whose ground metrics are the Mahalanobis distances.
For simplicity, we denote $\min_{\pi_t \in \mathcal U} \left< \pi_t, \Delta_t \right>$ as $L_\text{MW}^{(t)}$ as follows.

To calculate $L_\text{MW}^{(t)}$, we employ \emph{Sinkhorn iteration}~(Algorithm~\ref{algo:sinkhorn}), allowing us to accurately estimate the Wasserstein distance as well as its gradients \cite{cuturi2013sinkhorn}.
There are two benefits in the Sinkhorn iteration:
i) we can use GPUs as it combines some simple arithmetic operations;
ii) we can back-prop the iteration directly through auto-gradient techniques equipped in most modern deep learning libraries.

\begin{algorithm}[htbp]
  \caption{Computing Wasserstein distance by the Sinkhorn iteration. $\varepsilon>0$ is a small constant.}\label{algo:sinkhorn}
  \DontPrintSemicolon
  \KwData{$\Delta_t$}
  \nl $\vec{G}\leftarrow\exp(-\Delta_t/\varepsilon)$\tcp*{\footnotesize{Element-wise exp}}
  \nl $\vec{b}\leftarrow \onevec{N}$\tcp*{\footnotesize{Initialize dual variable}}
  \nl \While{Not converged}{
      \nl $\vec{a}\leftarrow \frac1N\onevec{N}/\vec{G}\vec{b}$\tcp*{\footnotesize{Element-wise division}}
      \nl $\vec{b}\leftarrow \frac1N\onevec{N}/\vec{G}^\top\vec{a}$
  }
  \nl \textbf{{return}} \, {$\langle\diag{\vec{a}}\vec{G}\,\diag{\vec{b}},\Delta_t \rangle$ as $L_\text{MW}^{(t)}$}\;
\end{algorithm}

To improve numerical stability, we may compute Algorithm~\ref{algo:sinkhorn} in the log domain~\cite{peyre2019computational}. 
Furthermore, we can compute Algorithm~\ref{algo:sinkhorn} in parallel over batch dimension.
At Line~4, we can use any type of stopping condition; here, we stop at $n_{it}$ iterations.
See \cite{cuturi2013sinkhorn,peyre2019computational} for details on the regularized Wasserstein distance.
In our training, $\varepsilon$ is set as 0.001 to suppress the approximation and to stably calculate the Mahalanobis-–Wasserstein loss, $L_\text{MW}$. We set the number of Sinkhorn iterations $n_{it}$ as 30 to obtain better coupling.

\begin{figure}[ht]
  \begin{center}
  \hspace*{-5mm}
      \includegraphics[width=0.95\hsize]{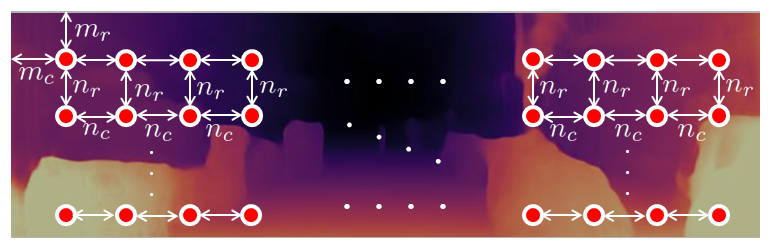}
  \end{center}
      \vspace*{-3mm}
	\caption{\small {\bf Sparse sampling of point clouds.} The red dots represent the positions of sampled point clouds in the image coordinates.}
  \label{f:sample}
  \vspace*{-3mm}
\end{figure}

\paragraph{Sparse Sampling}
To reduce the GPU memory usage of the Mahalanobis--Wasserstein loss, we uniformly sample the point clouds from the depth image, as shown in Fig.~\ref{f:sample}.
Here, $n_c$ and $n_r$ indicate the vertical and horizontal grid point intervals (in pixels), respectively.
$m_c$ and $m_r$ indicate random offsets of less than and equal to $n_c$ and $n_r$, respectively.
We cover the whole point clouds in the training by randomly choosing $m_c$ and $m_r$ in each training iteration.

\section{Metrics of Depth Evaluation}
In quantitative analysis, we evaluate the estimated depth using seven metrics.
``Abs-Rel,'' ``Sq Rel,'' ``RMSE,'' and ``RMSE log'' are calculated by the means of the following values in the entire test data. 
\begin{itemize}
  \item Abs-Rel\hspace{9mm}:\hspace{2mm} $|D_{gt} - \mu_{D_t}|/D_{gt}$
  \item Sq Rel\hspace{11mm}:\hspace{2mm} $(D_{gt} - \mu_{D_t})^2/D_{gt}$
  \item RMSE\hspace{11mm}:\hspace{2mm} $\sqrt{(D_{gt} - \mu_{D_t})^2}$
  \item RMSE log\hspace{6mm}:\hspace{2mm} $\sqrt{(\mbox{log}(D_{gt}) - \mbox{log}(\mu_{D_t}))^2}$
\end{itemize}
Here, $D_{gt}$ denotes the ground truth of the depth image. The remaining three metrics are the ratios that satisfy $\delta < \gamma$. $\delta$ is calculated as follows:
\begin{eqnarray}
    \delta = \mbox{max}(D_{gt}/\mu_{D_t}, \mu_{D_t}/D_{gt}).
    \label{eq:delta}
\end{eqnarray}
As with prior studies, we have three $\gamma$ values: 1.25, 1.25$^2$, and 1.25$^3$.

\section{Color map of Estimated Mean and Standard Deviation of Depth Image}
\begin{figure}[ht]
  \begin{center}
  %\hspace*{-5mm}
      \includegraphics[width=0.85\hsize]{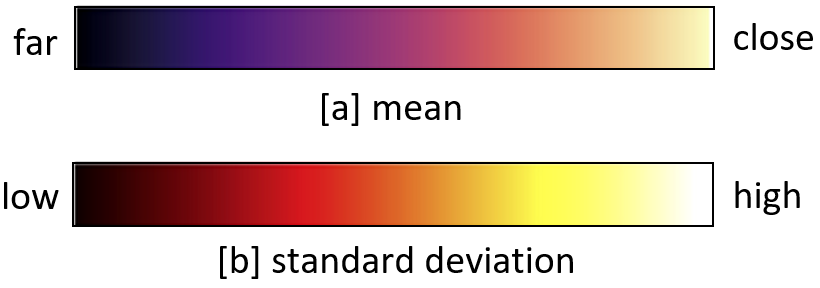}
  \end{center}
      \vspace*{-3mm}
  \caption{{\small {\bf Color map for mean and standard deviation of the depth image.} The relative standard deviation uses the same color map as the standard deviation.}}
  \label{f:colormap}
  \vspace*{-3mm}
\end{figure}

Fig.~\ref{f:colormap} is the color map of estimated mean and standard deviation of depth image. Fig.~\ref{f:colormap}[a] is for mean, and Fig.~\ref{f:colormap}[b] is for standard deviation.
\section{Outlier removal}
%
%In supplemental material, 
We show the actual values of Figure \ref{f:outlier} in Table~\ref{tab:ev_improve} and \ref{tab:ev_make3d}. Table~\ref{tab:ev_improve} is for KITTI, and Table~\ref{tab:ev_make3d} is for Make3D. We use bold to highlight the best results. Our method achieved the best results in almost all metrics and all removal percentages. 

\begin{table*}[t]
  \caption{{\bf Evaluation of depth estimation on an Eigen split of the KITTI dataset with outlier removal.} We display seven metrics from the estimated depth images of less than 80 m. For the left-most four metrics, smaller is better; for the right-most three metrics, higher is better. ``small'' denotes using 416$\times$128 image size. ``refine'' denotes using our proposed refinement in Section 5.3.3.}
  \vspace*{-3mm}
  \begin{center}
  \resizebox{1.99\columnwidth}{!}{
  \label{tab:ev_improve}
  \begin{tabular}{lcc|c|c|c|c||c|c|c} \hline
    Method & Image size& Percentage & \cellcolor[rgb]{0.9, 0.5, 0.5} Abs-Rel & \cellcolor[rgb]{0.9, 0.5, 0.5} Sq Rel & \cellcolor[rgb]{0.9, 0.5, 0.5} RMSE & \cellcolor[rgb]{0.9, 0.5, 0.5} RMSE log & \cellcolor[rgb]{0.5, 0.5, 0.9} $\delta < 1.25$ & \cellcolor[rgb]{0.5, 0.5, 0.9} $\delta < 1.25^2$ & \cellcolor[rgb]{0.5, 0.5, 0.9} $\delta < 1.25^3$ \\ \hline  
    monodetph2~\cite{godard2019digging,Poggi_CVPR_2020} & 640x192 & 0 $\%$ & 0.088 & 0.510 & 3.843 & 0.134 & 0.917 & 0.983 & 0.995 \\    
    & & 5 $\%$ & 0.087 & 0.496 & 3.795 & 0.131 & 0.920 & 0.984 & 0.996 \\
    & & 10 $\%$ & 0.085 & 0.487 & 3.766 & 0.129 & 0.922 & 0.985 & 0.996 \\
    & & 15 $\%$ & 0.084 & 0.479 & 3.739 & 0.128 & 0.924 & 0.985 & 0.996 \\    
    & & 20 $\%$ & 0.083 & 0.472 & 3.708 & 0.127 & 0.926 & 0.986 & 0.996 \\
    & & 30 $\%$ & 0.082 & 0.460 & 3.656 & 0.124 & 0.928 & 0.986 & 0.996 \\ \hline
    Poggi et al.~\cite{Poggi_CVPR_2020} & 640x192 & 0 $\%$ & 0.087 & 0.514 & 3.827 & 0.133 & \bf{0.920} & 0.983 & 0.995 \\    
    & & 5 $\%$ & 0.081 & 0.459 & 3.695 & 0.120 & 0.931 & \bf{0.988} & \bf{0.997} \\
    & & 10 $\%$ & 0.077 & 0.429 & 3.616 & 0.114 & 0.938 & \bf{0.990} & 0.997 \\
    & & 15 $\%$ & 0.074 & 0.403 & 3.534 & 0.109 & 0.943 & 0.991 & \bf{0.998} \\
    & & 20 $\%$ & 0.071 & 0.379 & 3.446 & 0.105 & 0.947 & 0.992 & 0.998 \\
    & & 30 $\%$ & 0.067 & 0.334 & 3.254 & 0.096 & 0.956 & \bf{0.995} & \bf{0.999} \\ \hline
    \bf{Our method(small)} & 416x128 & 0 $\%$ & 0.097 & 0.592 & 4.302 & 0.150 & 0.895 & 0.975 & 0.993 \\
    & & 5 $\%$ & 0.088 & 0.414 & 3.537 & 0.133 & 0.913 & 0.982 & 0.996 \\
    & & 10 $\%$ & 0.081 & 0.319 & 2.958 & 0.121 & 0.927 & 0.987 & 0.997 \\
    & & 15 $\%$ & 0.076 & 0.252 & 2.466 & 0.111 & 0.938 & 0.990 & \bf{0.998} \\    
    & & 20 $\%$ & 0.073 & 0.204 & 2.072 & 0.103 & 0.947 & 0.992 & \bf{0.999} \\
    & & 30 $\%$ & 0.067 & 0.143 & 1.524 & 0.092 & 0.960 & \bf{0.995} & \bf{0.999} \\ \hline
    \bf{Our method(small) + refine} & 416x128 & 0 $\%$ & 0.093 & 0.552 & 4.171 & 0.145 & 0.902 & 0.977 & 0.994 \\
    & & 5 $\%$ & 0.085 & 0.389 & 3.442 & 0.129 & 0.918 & 0.984 & 0.997 \\
    & & 10 $\%$ & 0.079 & 0.301 & 2.883 & 0.118 & 0.931 & 0.988 & \bf{0.998} \\
    & & 15 $\%$ & 0.074 & 0.238 & 2.404 & 0.108 & 0.942 & 0.991 & \bf{0.998} \\    
    & & 20 $\%$ & 0.071 & 0.193 & 2.018 & 0.101 & 0.951 & 0.993 & \bf{0.999} \\
    & & 30 $\%$ & 0.066 & 0.136 & 1.481 & 0.090 & 0.962 & \bf{0.995} & \bf{0.999} \\ \hline
    \bf{Our method} & 640x192 & 0 $\%$ & 0.088 & 0.488 & 3.833 & 0.134 & 0.915 & 0.983 & \bf{0.996} \\
    & & 5 $\%$ & 0.080 & 0.324 & 2.938 & 0.121 & 0.929 & 0.987 & \bf{0.997} \\
    & & 10 $\%$ & 0.075 & 0.249 & 2.364 & 0.113 & 0.940 & \bf{0.990} & \bf{0.998} \\
    & & 15 $\%$ & 0.071 & 0.198 & 1.947 & 0.105 & 0.949 & \bf{0.992} & \bf{0.998} \\    
    & & 20 $\%$ & 0.068 & 0.162 & 1.649 & 0.098 & 0.956 & 0.993 & \bf{0.999} \\
    & & 30 $\%$ & 0.064 & 0.123 & 1.307 & 0.090 & 0.964 & \bf{0.995} & \bf{0.999} \\ \hline
    \bf{Our method + refine} & 640x192 & 0 $\%$ & \bf{0.085} & \bf{0.455} & \bf{3.707} & \bf{0.130} & \bf{0.920} & \bf{0.984} & \bf{0.996} \\ 
    & & 5 $\%$ & \bf{0.078} & \bf{0.304} & \bf{2.841} & \bf{0.119} & \bf{0.933} & \bf{0.988} & \bf{0.997} \\
    & & 10 $\%$ & \bf{0.073} & \bf{0.235} & \bf{2.294} & \bf{0.111} & \bf{0.943} & \bf{0.990} & \bf{0.998} \\
    & & 15 $\%$ & \bf{0.070} & \bf{0.188} & \bf{1.894} & \bf{0.103} & \bf{0.951} & \bf{0.992} & \bf{0.998} \\    
    & & 20 $\%$ & \bf{0.067} & \bf{0.155} & \bf{1.607} & \bf{0.097} & \bf{0.957} & \bf{0.994} & \bf{0.999} \\
    & & 30 $\%$ & \bf{0.063} & \bf{0.118} & \bf{1.278} & \bf{0.089} & \bf{0.965} & \bf{0.995} & \bf{0.999} \\ \hline    
  \end{tabular}
  }
  \end{center}
  %\vspace{-0.5em}
\end{table*}
\begin{table*}[t]
  \caption{{\bf Evaluation of depth estimation on the Make3D dataset with outlier removal.} We display seven metrics from the estimated depth images of less than 70 m. For the left-most four metrics, smaller is better; for the right-most three metrics, higher is better. ``small'' denotes using 416$\times$128 image size.}
  \vspace*{-3mm}
  \begin{center}
  \resizebox{1.99\columnwidth}{!}{
  \label{tab:ev_make3d}
  \begin{tabular}{lcc|c|c|c|c||c|c|c} \hline
    Method & Image size& Percentage & \cellcolor[rgb]{0.9, 0.5, 0.5} Abs-Rel & \cellcolor[rgb]{0.9, 0.5, 0.5} Sq Rel & \cellcolor[rgb]{0.9, 0.5, 0.5} RMSE & \cellcolor[rgb]{0.9, 0.5, 0.5} RMSE log & \cellcolor[rgb]{0.5, 0.5, 0.9} $\delta < 1.25$ & \cellcolor[rgb]{0.5, 0.5, 0.9} $\delta < 1.25^2$ & \cellcolor[rgb]{0.5, 0.5, 0.9} $\delta < 1.25^3$ \\ \hline  
    monodetph2~\cite{godard2019digging,Poggi_CVPR_2020} & 640x192 & 0 $\%$ & 0.287 & 3.205 & 6.946 & 0.326 & 0.621 & 0.846 & 0.938 \\    
    & & 5 $\%$ & 0.283 & 3.117 & 6.890 & 0.323 & 0.622 & 0.848 & 0.939 \\
    & & 10 $\%$ & 0.278 & 3.001 & 6.856 & 0.320 & 0.624 & 0.850 & 0.940 \\
    & & 15 $\%$ & 0.275 & 2.937 & 6.836 & 0.318 & 0.624 & 0.850 & 0.941 \\    
    & & 20 $\%$ & 0.273 & 2.887 & 6.813 & 0.317 & 0.624 & 0.851 & 0.942 \\
    & & 30 $\%$ & 0.271 & 2.834 & 6.762 & 0.315 & 0.626 & 0.852 & 0.943 \\ \hline  
    Poggi et al.~\cite{Poggi_CVPR_2020} & 640x192 & 0 $\%$ & \bf{0.282} & 3.073 & \bf{6.819} & \bf{0.323} & 0.623 & \bf{0.848} & \bf{0.943} \\    
    & & 5 $\%$ & 0.279 & 3.055 & 6.797 & 0.320 & 0.627 & 0.850 & 0.944 \\
    & & 10 $\%$ & 0.278 & 3.056 & 6.797 & 0.319 & 0.630 & 0.849 & 0.943 \\
    & & 15 $\%$ & 0.278 & 3.074 & 6.799 & 0.317 & 0.632 & 0.849 & 0.943 \\
    & & 20 $\%$ & 0.278 & 3.104 & 6.797 & 0.316 & 0.635 & 0.849 & 0.943 \\
    & & 30 $\%$ & 0.280 & 3.200 & 6.803 & 0.316 & 0.639 & 0.849 & 0.941 \\ \hline
    \bf{Our method(small)} & 416x128 & 0 $\%$ & 0.285 & \bf{2.880} & 6.915 & 0.333 & 0.591 & 0.843 & 0.935 \\
    & & 5 $\%$ & 0.269 & \bf{2.421} & 6.424 & 0.319 & 0.603 & 0.854 & 0.941 \\
    & & 10 $\%$ & 0.258 & \bf{2.187} & 6.095 & 0.310 & 0.613 & 0.861 & 0.945 \\
    & & 15 $\%$ & 0.252 & \bf{2.032} & 5.818 & 0.303 & 0.622 & 0.866 & 0.948 \\
    & & 20 $\%$ & 0.246 & \bf{1.903} & 5.557 & 0.296 & 0.629 & 0.872 & 0.950 \\    
    & & 30 $\%$ & 0.237 & \bf{1.728} & 5.082 & 0.286 & 0.643 & 0.879 & 0.953 \\ \hline
    \bf{Our method} & 640x192 & 0 $\%$ & 0.283 & 3.072 & 6.902 & 0.327 & \bf{0.626} & 0.843 & 0.934 \\
    & & 5 $\%$ & \bf{0.264} & 2.584 & \bf{6.321} & \bf{0.311} & \bf{0.641} & \bf{0.855} & \bf{0.942} \\
    & & 10 $\%$ & \bf{0.251} & 2.296 & \bf{5.914} & \bf{0.299} & \bf{0.654} & \bf{0.863} & \bf{0.947} \\
    & & 15 $\%$ & \bf{0.242} & 2.117 & \bf{5.597} & \bf{0.292} & \bf{0.664} & \bf{0.870} & \bf{0.951} \\    
    & & 20 $\%$ & \bf{0.235} & 1.971 & \bf{5.309} & \bf{0.285} & \bf{0.674} & \bf{0.875} & \bf{0.954} \\
    & & 30 $\%$ & \bf{0.223} & 1.755 & \bf{4.907} & \bf{0.275} & \bf{0.687} & \bf{0.881} & \bf{0.958} \\ \hline
  \end{tabular}
  }
  \end{center}
  %\vspace{-0.5em}
\end{table*}
\section{Parameter Study of Refinement}
\blue{In Sec.~\ref{sec:refinement-using-estimated-std},} we propose a computationally light refinement algorithm using the estimated standard deviation. We herein show the results with the variance of the hyperparameters $N_k$ and $\alpha$. In addition, we measured the additional calculation time by implementing our method on a NVIDIA RTX 2080 Ti. Table~\ref{tab:ev_refinement} shows that our method can perform better when the sampling number $N_k$ is increased. However, the performance improvement is almost saturated at $N_k = 10$. Moreover, the additional calculation time of $N_k = 10$ is only 2.7 ms, which is much faster than the camera frame rate of 100.0 ms. In contrast to $N_k$, $\alpha$ varies slightly depending on the metric to derive the best result, but it seems that $\alpha = 0.2$ will be the best value. Hence, we selected $N_k = 10$ and $\alpha = 0.2$ in the experiment.

In addition, Table~\ref{tab:ev_large} includes the baselines that are not monodepth2 base. 
From Table~\ref{tab:ev_large}, PackNet-Sfm~\cite{guizilini2019packnet} and UnRectDepthNet~[\textcolor{green}{51}] are competitive. And, they are stronger than our method in the evaluation with improved ground truth.
%In the evaluation using the original ground truth, our method with refinement takes the best results in all metrics. 
%However, PackNet-Sfm~\cite{guizilini2019packnet} is stronger than our method in the evaluation with improved ground truth. 
Although we employ the same network structure and the same image reconstruction loss as monodepth2~\cite{godard2019digging} and Poggi et al.\cite{Poggi_CVPR_2020} to perform a comparative evaluation of the estimated distribution, the network structure and the loss need to be modified for better mean estimation performance. 
Our focus is the estimation of the depth image distribution to predict the reliability level, unlike the baselines, except that of Poggi et al.\cite{Poggi_CVPR_2020}.
\begin{table*}[ht]
  \caption{{\bf Parameter study of our refinement algorithm using estimated standard deviation.} We show the results using the original ground truth~\cite{eigen2014depth} and the improved one~\cite{uhrig2017sparsity}. In the top part, we change the sampling number $N_k$ under a fixed range of $\alpha$ = 0.2 in each tables. By contrast, we vary $\alpha$ with fixed $N_k$ = 10 in the bottom part.}
  \vspace*{-3mm}
  \begin{center}
  \resizebox{1.99\columnwidth}{!}{
  \label{tab:ev_refinement}
  \begin{tabular}{c|lc|c|c|c|c||c|c|c||c} \hline
    & $\alpha$ & $N_{k}$ & \cellcolor[rgb]{0.9, 0.5, 0.5}, Rel & \cellcolor[rgb]{0.9, 0.5, 0.5} Sq Rel & \cellcolor[rgb]{0.9, 0.5, 0.5} RMSE & \cellcolor[rgb]{0.9, 0.5, 0.5} RMSE log & \cellcolor[rgb]{0.5, 0.5, 0.9} $\delta < 1.25$ & \cellcolor[rgb]{0.5, 0.5, 0.9} $\delta < 1.25^2$ & \cellcolor[rgb]{0.5, 0.5, 0.9} $\delta < 1.25^3$ & \cellcolor[rgb]{0.9, 0.5, 0.5} add. calc. Time [ms] \\ \hline   
    \multicolumn{1}{c|}{\multirow{10}*{\rotatebox[origin=c]{90}{Original~\cite{eigen2014depth}}}} & 0.2 & 3 & \bf{0.110} & 0.797 & 4.642 & 0.188 & 0.880 & \bf{0.961} & \bf{0.982} & 1.825 \\ % 3.20
    & 0.2 & 5 & \bf{0.110} & 0.790 & 4.614 & \bf{0.187} & \bf{0.881} & \bf{0.961} & \bf{0.982} & 2.410 \\ % 3.95
    & 0.2 & 10 & \bf{0.110} & 0.786 & 4.597 & \bf{0.187} & \bf{0.881} & \bf{0.961} & \bf{0.982} & 2.725 \\ % 5.06
    & 0.2 & 20 & 0.111 & 0.784 & 4.591 & \bf{0.187} & \bf{0.881} & \bf{0.961} & \bf{0.982} & 3.894 \\ % 9.79
    & 0.2 & 100 & 0.111 & 0.783 & 4.586 & 0.188 & \bf{0.881} & \bf{0.961} & \bf{0.982} & 12.301 \\ \cline{2-11} %126.07 
    & 0.05 & 10 & 0.111 & 0.802 & 4.659 & 0.188 & 0.879 & \bf{0.961} & \bf{0.982} & $-$ \\  
    & 0.1 & 10 & \bf{0.110} & 0.795 & 4.635 & \bf{0.187} & 0.880 & \bf{0.961} & \bf{0.982} & $-$ \\      
    & 0.2 & 10 & \bf{0.110} & 0.786 & 4.597 & \bf{0.187} & \bf{0.881} & \bf{0.961} & \bf{0.982} & $-$ \\    
    & 0.3 & 10 & 0.112 & \bf{0.781} & \bf{4.575} & 0.188 & 0.880 & \bf{0.961} & \bf{0.982} & $-$ \\ 
    & 1.0 & 10 & 1.34 & 0.869 & 4.727 & 0.212 & 0.824 & 0.953 & 0.981 & $-$ \\ \hline\hline  
    \multicolumn{1}{c|}{\multirow{9}*{\rotatebox[origin=c]{90}{Improved~\cite{uhrig2017sparsity}}}} & 0.2 & 3 & 0.086 & 0.471 & 3.779 & 0.132 & 0.918 & \bf{0.984} & \bf{0.996} & 1.825 \\    
    & 0.2 & 5 & \bf{0.085} & 0.461 & 3.735 & 0.131 & 0.920 & \bf{0.984} & \bf{0.996} & 2.410 \\     
    & 0.2 & 10 & \bf{0.085} & 0.455 & 3.707 & \bf{0.130} & 0.920 & \bf{0.984} & \bf{0.996} & 2.725 \\ 
    & 0.2 & 20 & \bf{0.085} & 0.453 & 3.694 & \bf{0.130} & \bf{0.921} & \bf{0.984} & \bf{0.996} & 3.894 \\ 
    & 0.2 & 100 & \bf{0.085} & \bf{0.451} & 3.685 & \bf{0.130} & \bf{0.921} & \bf{0.984} & \bf{0.996} & 12.301 \\ \cline{2-11}    
    & 0.05 & 10 & 0.086 & 0.477 & 3.804 & 0.132 & 0.917 & 0.983 & \bf{0.996} & $-$ \\  
    & 0.1 & 10 & \bf{0.085} & 0.468 & 3.767 & 0.131 & 0.919 & \bf{0.984} & \bf{0.996} & $-$ \\      
    & 0.2 & 10 & \bf{0.085} & 0.455 & 3.707 & \bf{0.130} & 0.920 & \bf{0.984} & \bf{0.996} & $-$ \\    
    & 0.3 & 10 & 0.086 & 0.447 & \bf{3.663} & 0.131 & \bf{0.921} & \bf{0.984} & \bf{0.996} & $-$ \\ 
    & 1.0 & 10 & 1.038 & 0.491 & 3.701 & 0.154 & 0.876 & 0.979 & 0.995 & $-$ \\ \hline      
  \end{tabular}
  }
  \end{center}
  %\vspace{-0.5em}
\end{table*}
\begin{table*}[t]
  \caption{{\small {\bf Evaluation of depth estimation by self-supervised mono-supervision on an Eigen split of the KITTI dataset.} We display seven metrics from the {\bf estimated depth images of less than 80 m}. For the left-most four metrics, smaller is better; for the right-most three metrics, higher is better. Here, ``refine'' denotes using our proposed refinement in Section 5.3.3. ``flip'' denotes the post process using horizontally flipped image shown in Section 5.3.1. ``$\dagger$'' denotes using the model trained by ourselves because monodepth2 does not open the model for a 416$\times$128 image size.}}
  %\vspace*{-3mm}
  \begin{center}
  \resizebox{1.99\columnwidth}{!}{
  \label{tab:ev_large}
  \begin{tabular}{c|c|lc|c|c|c|c||c|c|c} \hline
    & \multicolumn{2}{|l}{Method} & Image size & \cellcolor[rgb]{0.9, 0.5, 0.5} Abs-Rel & \cellcolor[rgb]{0.9, 0.5, 0.5} Sq Rel & \cellcolor[rgb]{0.9, 0.5, 0.5} RMSE & \cellcolor[rgb]{0.9, 0.5, 0.5} RMSE log & \cellcolor[rgb]{0.5, 0.5, 0.9} $\delta < 1.25$ & \cellcolor[rgb]{0.5, 0.5, 0.9} $\delta < 1.25^2$ & \cellcolor[rgb]{0.5, 0.5, 0.9} $\delta < 1.25^3$ \\ \hline 
    \multicolumn{1}{c|}{\multirow{23}*{\rotatebox[origin=c]{90}{Original~\cite{eigen2014depth}}}} & \multicolumn{1}{c|}{\multirow{14}*{\rotatebox[origin=c]{90}{$-$}}} & Zhou et al.~\cite{zhou2017unsupervised} & 416x128 & 0.208 & 1.768 & 6.856 & 0.283 & 0.678 & 0.885 & 0.957 \\ 
    & & Yang et al.~\cite{yang2017unsupervised} & 416x128 & 0.182 & 1.481 & 6.501 & 0.267 & 0.725 & 0.906 & 0.963 \\
    & & vid2depth~\cite{mahjourian2018unsupervised} & 416x128 & 0.163 & 1.240 & 6.220 & 0.250 & 0.762 & 0.916 & 0.968 \\
    & & LEGO~\cite{yang2018lego} & 416x128 & 0.162 & 1.352 & 6.276 & 0.252 & 0.783 & 0.921 & 0.969 \\
    & & GeoNet~\cite{yin2018geonet} & 416x128 & 0.155 & 1.296 & 5.857 & 0.233 & 0.793 & 0.931 & 0.973 \\
    & & Fei et al.~\cite{fei2019geo} & 416x128 & 0.142 & 1.124 & 5.611 & 0.223 & 0.813 & 0.938 & 0.975 \\
    & & DDVO~\cite{wang2018learning} & 416x128 & 0.151 & 1.257 & 5.583 & 0.228 & 0.810 & 0.936 & 0.974 \\
    & & WBAF~\cite{wbaf2020} & 416x128 & 0.135 & 0.992 & 5.288 & 0.211 & 0.831 & 0.942 & 0.976 \\    
    & & Yang et al.~\cite{yang2018every} & 416x128 & 0.131 & 1.254 & 6.117 & 0.220 & 0.826 & 0.931 & 0.973 \\
    & & Casser et al.~\cite{casser2019depth} & 416x128 & 0.141 & 1.026 & 5.291 & 0.2153 & 0.8160 & 0.9452 & 0.9791 \\
    & & Gordon et al.~\cite{gordon2019depth} & 416x128 & 0.128 & 0.959 & 5.23 & 0.212 & 0.845 & 0.947 & 0.976 \\
    & & PRGB-D Refined~\cite{tiwari2020pseudo} & 640x192 & 0.113 & 0.793 & 4.655 & 0.188 & 0.874 & 0.960 & 0.983 \\
    & & PackNet-SfM~\cite{guizilini2019packnet} & 640x192 & 0.111 & 0.785 & 4.601 & 0.189 & 0.878 & 0.960 & 0.982 \\
    & & UnRectDepthNet~\cite{kumar2020unrectdepthnet} & 640x192 & 0.107 & 0.721 & 4.564 & 0.178 & 0.894 & 0.971 & 0.986 \\ \cline{2-11}    
    & \multicolumn{1}{c|}{\multirow{11}*{\rotatebox[origin=c]{90}{monodepth2 base}}} & monodepth2~\cite{godard2019digging} & 416x128 & 0.128 & 1.087 & 5.171 & 0.204 & 0.855 & 0.953 & 0.978 \\
    & & monodepth2 + flip~\cite{godard2019digging} $\dagger$ & 416x128 & 0.128 & 1.108 & 5.081 & 0.203 & 0.857 & 0.954 & 0.979 \\
    & & PLG-IN~\cite{hirose2020plg} & 416x128 & 0.123 & 0.920 & 4.990 & 0.201 & 0.858 & 0.953 & 0.980 \\
    & & \bf{Our method} & 416x128 & 0.121 & 0.883 & 4.936 & 0.198 & 0.860 & 0.954 & 0.980 \\
    & & \bf{Our method + refine} & 416x128 & \underline{0.118} & \underline{0.848} & \underline{4.845} & \underline{0.196} & \underline{0.865} & \underline{0.956} & \underline{0.981} \\    
    & & monodepth2~\cite{godard2019digging} & 640x192 & 0.115 & 0.903 & 4.863 & 0.193 & 0.877 & 0.959 & 0.981 \\
    & & monodepth2 + flip~\cite{godard2019digging} & 640x192 & 0.112 & 0.851 & 4.754 & 0.190 & 0.881 & 0.960 & 0.981 \\    
    & & PLG-IN~\cite{hirose2020plg} & 640x192 & 0.114 & 0.813 & 4.705 & 0.191 & 0.874 & 0.959 & 0.981 \\
    & & Poggi et al.~\cite{Poggi_CVPR_2020}  & 640x192 & 0.111 & 0.863 & 4.756 & 0.188 & \bf{0.881} & \bf{0.961} & \bf{0.982} \\        
    %& Our method: & & & & & & & & \\
    & & \bf{Our method} & 640x192 & 0.112 & 0.813 & 4.688 & 0.189 & 0.878 & 0.960 & \bf{0.982} \\   
    & & \bf{Our method + refine} & 640x192 & \bf{0.110} & \bf{0.785} & \bf{4.597} & \bf{0.187} & \bf{0.881} & \bf{0.961} & \bf{0.982} \\ \hline\hline  
    \multicolumn{1}{c|}{\multirow{11}*{\rotatebox[origin=c]{90}{Improved~\cite{uhrig2017sparsity}}}} & \multicolumn{1}{c|}{\multirow{3}*{\rotatebox[origin=c]{90}{$-$}}} & EPC++~\cite{fu2018deep} & 640x192 & 0.120 & 0.789 & 4.755 & 0.177 & 0.856 & 0.961 & 0.987 \\ 
    & & PackNet-SfM~\cite{guizilini2019packnet} & 640x192 & 0.078 & 0.420 & 3.485 & 0.121 & 0.931 & 0.986 & 0.996 \\ 
    & & UnRectDepthNet~\cite{kumar2020unrectdepthnet} & 640x192 & 0.081 & 0.414 & 3.412 & 0.117 & 0.926 & 0.987 & 0.996 \\ \cline{2-11}     
    & \multicolumn{1}{c|}{\multirow{9}*{\rotatebox[origin=c]{90}{monodepth2 base}}} & monodepth2~\cite{godard2019digging} & 416x128 & 0.108 & 0.767 & 4.343 & 0.156 & 0.890 & 0.974 & 0.991 \\    
    & & monodepth2 + flip~\cite{godard2019digging} $\dagger$ & 416x128 & 0.104 & 0.716 & 4.233 & 0.152 & 0.894 & 0.975 & 0.992 \\    
    & & \bf{Our method} & 416x128 & 0.097 & 0.592 & 4.302 & 0.150 & 0.895 & 0.975 & 0.993 \\
    & & \bf{Our method + refine} & 416x128 & \underline{0.093} & \underline{0.552} & \underline{4.171} & \underline{0.145} & \underline{0.902} & \underline{0.977} & \underline{0.994} \\    
    & & monodepth2~\cite{godard2019digging} & 640x192 & 0.090 & 0.545 & 3.942 & 0.137 & 0.914 & 0.983 & 0.995 \\
    & & monodepth2 + flip~\cite{godard2019digging} & 640x192 & 0.087 & 0.470 & 3.800 & 0.132 & 0.917 & 0.984 & 0.996 \\
    & & Poggi et al.~\cite{Poggi_CVPR_2020}  & 640x192 & 0.087 & 0.514 & 3.827 & 0.133 & \bf{0.920} & 0.983 & 0.995 \\      
    & & \bf{Our method} & 640x192 & 0.088 & 0.488 & 3.833 & 0.134 & 0.915 & 0.983 & \bf{0.996} \\
    & & \bf{Our method + refine} & 640x192 & \bf{0.085} & \bf{0.455} & \bf{3.707} & \bf{0.130} & \bf{0.920} & \bf{0.984} & \bf{0.996} \\ \hline
  \end{tabular}
  }
\end{center}
  \vspace{-5mm}
\end{table*}
\section{Ablation Study of Post Processes}
Table~\ref{tab:ev_pp} shows the ablation study of post processes. In Table~\ref{tab:ev_pp}, we apply our refinement using estimated standard deviation(``refine''), the method with flipped image~\cite{godard2019digging}(``flip''), or both of them. The baseline post process ``flip'' is explained in Section 5.3.1. When we apply both, we estimate flipped mean and flipped standard deviation of the depth image from the flipped image. Subsequently, we calculate their mean by flipping them again to calculate ``refine.'' From Table~\ref{tab:ev_pp}, we can confirm that ``refine'' is more effective than ``flip.'' In addition, the best results are obtained when both are applied.

\begin{table*}[ht]
  \caption{{\bf Ablation study of post processes.} ``refine'' denotes our method using estimated standard deviation. ``flip'' denotes the approach using flipped image~\cite{godard2019digging}. We apply these post processes to our method with a 640$\times$192 image size to evaluate the performance.}
  \vspace*{0mm}
  \begin{center}
  \resizebox{1.8\columnwidth}{!}{
  \label{tab:ev_pp}
  \begin{tabular}{c|cc|c|c|c|c||c|c|c} \hline
    & refine & flip~\cite{godard2019digging} & \cellcolor[rgb]{0.9, 0.5, 0.5} Abs-Rel & \cellcolor[rgb]{0.9, 0.5, 0.5} Sq Rel & \cellcolor[rgb]{0.9, 0.5, 0.5} RMSE & \cellcolor[rgb]{0.9, 0.5, 0.5} RMSE log & \cellcolor[rgb]{0.5, 0.5, 0.9} $\delta < 1.25$ & \cellcolor[rgb]{0.5, 0.5, 0.9} $\delta < 1.25^2$ & \cellcolor[rgb]{0.5, 0.5, 0.9} $\delta < 1.25^3$ \\ \hline   
    \multicolumn{1}{c|}{\multirow{4}*{\rotatebox[origin=c]{90}{Orig.~\cite{eigen2014depth}}}} & & & 0.112 & 0.813 & 4.688 & 0.189 & 0.878 & 0.960 & 0.982 \\ %1.825 
    & & \checkmark & 0.111 & 0.784 & 4.638 & 0.187 & 0.879 & 0.961 & 0.982 \\ %2.725     
    & \checkmark & & 0.110 & 0.785 & 4.597 & 0.187 & 0.881 & 0.961 & 0.982 \\ %2.410  
    & \checkmark & \checkmark & \bf{0.109} & \bf{0.763} & \bf{4.542} & \bf{0.185} & \bf{0.883} & \bf{0.962} & \bf{0.983} \\ \hline\hline  
    \multicolumn{1}{c|}{\multirow{4}*{\rotatebox[origin=c]{90}{Imp.~\cite{uhrig2017sparsity}}}} & & & 0.088 & 0.488 & 3.833 & 0.134 & 0.915 & 0.983 & \bf{0.996} \\    
    & & \checkmark & 0.087 & 0.470 & 3.800 & 0.132 & 0.917 & 0.984 & \bf{0.996} \\     
    & \checkmark & & 0.085 & 0.455 & 3.707 & 0.130 & 0.920 & 0.984 & \bf{0.996} \\     
    & \checkmark & \checkmark & \bf{0.084} & \bf{0.440} & \bf{3.654} & \bf{0.129} & \bf{0.922} & \bf{0.985} & \bf{0.996} \\ \hline
  \end{tabular}
  }
  \end{center}
  %\vspace{-0.5em}
\end{table*}
%
%\clearpage
%\clearpage
\section{Additional qualitative analysis}
\begin{figure*}[t]
  \begin{center}
  %\hspace*{-5mm}
      \includegraphics[width=0.92\hsize]{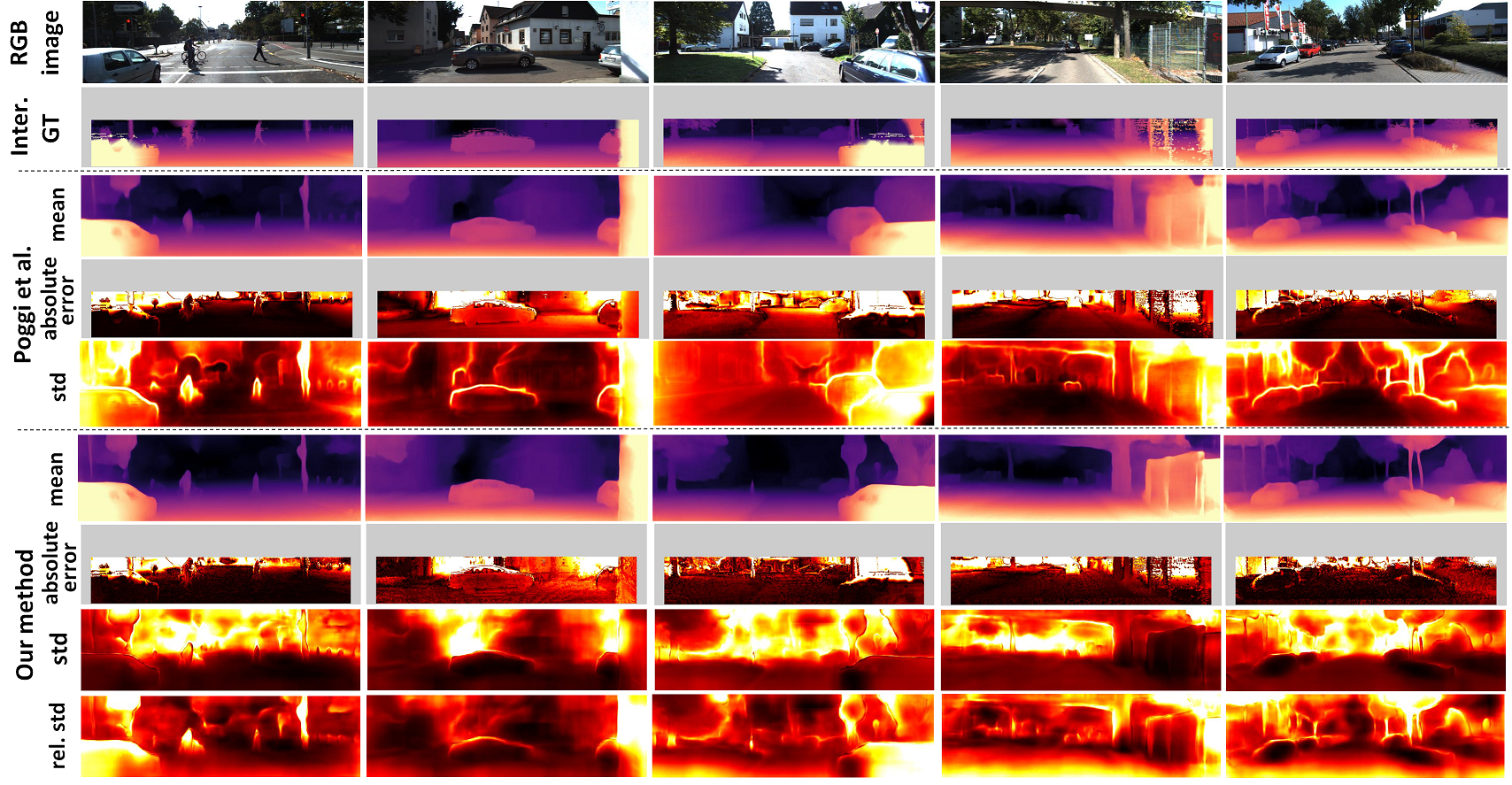}
  \end{center}
      \vspace*{-3mm}
  \caption{{\small {\bf Visualization of the proposed variational depth estimation in KITTI dataset.} From the top, we display the RGB image, interpolated ground truth depth image, estimated mean, absolute error using the interpolated ground truth, and standard deviation of depth image by Poggi et al.~\cite{Poggi_CVPR_2020} and estimated mean, absolute error using the interpolated ground truth, and standard deviation of the depth image by our method. At the bottom, we show the relative standard deviation when using our method to highlight low-reliability pixels except for the farther areas. Our standard deviation is visually close to the absolute error.}}
  \label{f:depth_n}
  \vspace*{-3mm}
\end{figure*}

\begin{figure*}[t]
  \begin{center}
  %\hspace*{-5mm}
      \includegraphics[width=0.92\hsize]{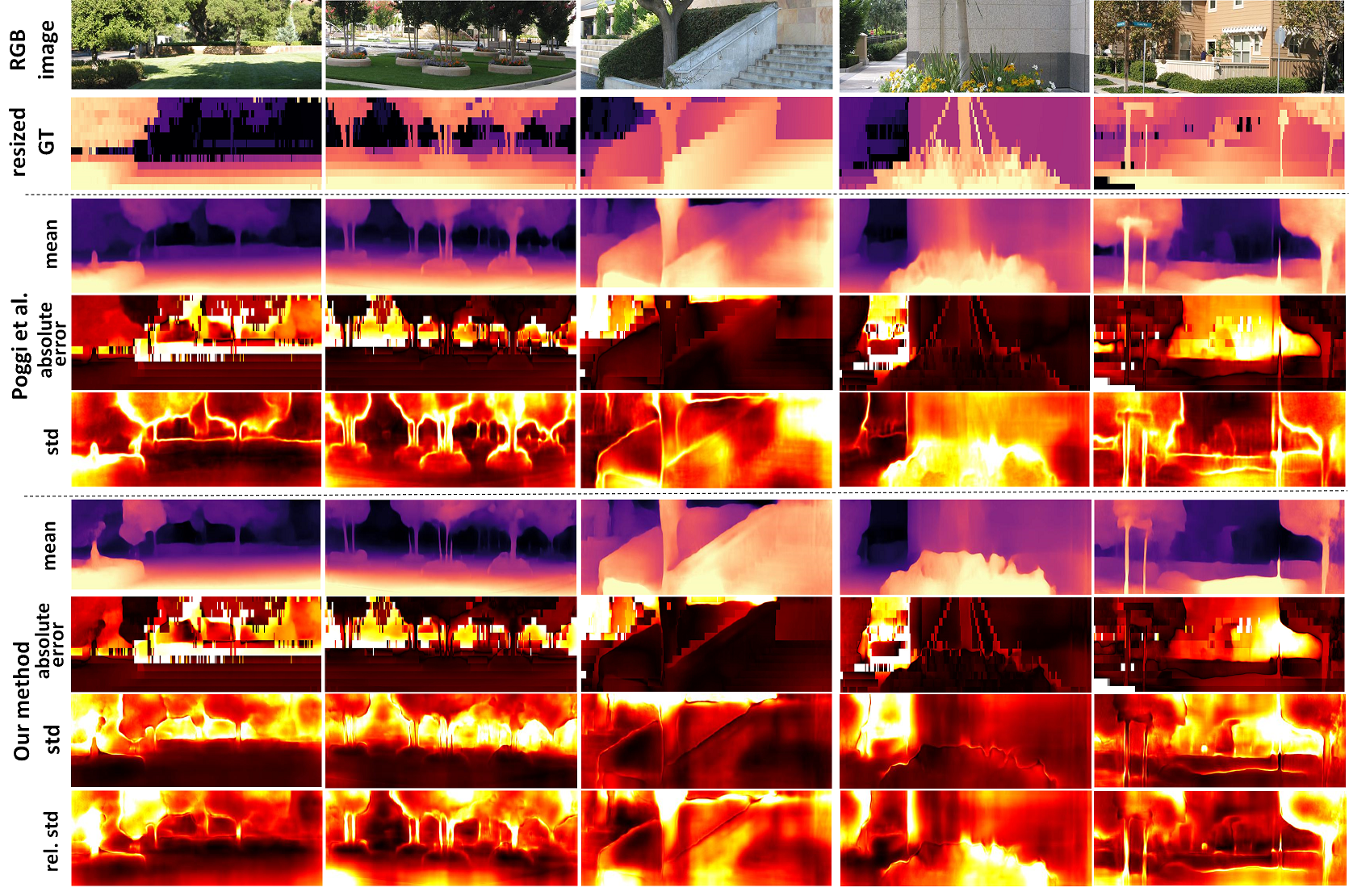}
  \end{center}
      \vspace*{-3mm}
  \caption{{\small {\bf Visualization of the proposed variational depth estimation in the Make3D dataset.} The configuration is the same as that shown in Fig.\ref{f:depth_n}.}}
  \label{f:depth_n_make3d}
  \vspace*{-3mm}
\end{figure*}

Figures~\ref{f:depth_n} and \ref{f:depth_n_make3d} are visualizations of the proposed variational depth estimation in KITTI and Make3D as the additional qualitative analysis. In addition to the estimated mean and standard deviation of the depth image, we demonstrate the interpolated ground truth and the absolute error between its interpolated ground truth and the estimated mean depth to evaluate the estimated standard deviation. In Fig.~\ref{f:depth_n}, the gray color denotes the area wherein there is no ground truth depth. Our estimated standard deviation is visually close to the absolute error, indicating that our method can correctly estimate the reliability of depth estimation. In contrast to our method, the gap between the absolute error and estimated standard deviation as done by Poggi et al. is large at the dynamic objects, the edges between different objects and distant area. As the ground truth of Make3D has low resolution, the calculated absolute error is discontinuous. However, we can confirm that our standard deviation is mostly consistent with the absolute error,  visually. In contrast to our method, Poggi et al. showed higher standard deviation at edges between different objects and close objects and lower values at the distant area, which are inconsistent with the absolute error.
%
%
%\vspace{5mm}
%\noindent
%\textbf{Reference}
%
%\noindent
%{\small [51] Varun Ravi Kumar, Senthil Yogamani, Markus Bach, Christian Witt, Stefan Milz, and Patrick Mader, ``UnRectDepthNet: Self-Supervised Monocular Depth Estimation using a Generic Framework for Handling Common Camera Distortion Models'', In Proceedings of IEEE/RSJ International Conference of Intelligent Robots and Systems, pages 8177--8183, 2020}

\end{document}